\PassOptionsToPackage{dvipsnames,table}{xcolor}
\documentclass[]{bytedance_seed}
\usepackage{minitoc}
\usepackage[toc,page,header]{appendix}
\PassOptionsToPackage{numbers, compress}{natbib}

\usepackage{cleveref} 
\usepackage{booktabs}
\usepackage{xspace}
\usepackage{graphicx}
\usepackage{multirow}
\usepackage{multicol}
\usepackage{amssymb}
\usepackage{amsmath}

\newcommand\mycheckmark{{\color{ForestGreen}\textbf{\checkmark}}}
\newcommand{\newcheckmark}{{\color{gray}\checkmark}}
\newcommand\mycrossmark{{\color{Red}\textbf{$\times$}}}

\newcommand{\framework}{ByteMorph\xspace}
\newcommand{\model}{ByteMorpher\xspace}
\newcommand{\dataset}{ByteMorph-6M\xspace}
\newcommand{\benchmark}{ByteMorph-Bench\xspace}

\title{ByteMorph: Benchmarking Instruction-Guided Image Editing with Non-Rigid Motions}

\contribution{Di Chang$^{*1,2}$, Mingdeng Cao$^{*1,3}$, Yichun Shi$^{^1}$, Bo Liu$^{1,4}$, Shengqu Cai$^{1,5}$, Shijie Zhou$^6$\\  Weilin Huang$^{1}$, Gordon Wetzstein$^5$, Mohammad Soleymani$^2$, Peng Wang$^{^1}$\\$^*$Equal Contribution}

\affiliation[1]{ByteDance Seed\quad $^2$University of Southern California\quad $^3$University of Tokyo}
\affiliation[4]{University of California Berkeley\quad $^5$Stanford University\quad $^6$University of California Los Angeles}

\abstract{
Editing images with instructions to reflect non-rigid motions—camera viewpoint shifts, object deformations, human articulations, and complex interactions—poses a challenging yet underexplored problem in computer vision. Existing approaches and datasets predominantly focus on static scenes or rigid transformations, limiting their capacity to handle expressive edits involving dynamic motion. To address this gap, we introduce \framework, a comprehensive framework for instruction-based image editing with an emphasis on non-rigid motions. \framework comprises a large-scale dataset, \dataset, and a strong baseline model built upon the Diffusion Transformer (DiT), named \model. \dataset includes over 6 million high-resolution image editing pairs for training, along with a carefully curated evaluation benchmark \benchmark. Both capture a wide variety of non-rigid motion types across diverse environments, human figures, and object categories. The dataset is constructed using motion-guided data generation, layered compositing techniques, and automated captioning to ensure diversity, realism, and semantic coherence. We further conduct a comprehensive evaluation of recent instruction-based image editing methods from both academic and commercial domains.

}

\date{June 1, 2025}
\correspondence{\email{dichang@usc.edu}, \email{yichun.shi@bytedance.com}, \email{peng.wang@bytedance.com}}

\checkdata[Online Demo]{\url{https://huggingface.co/spaces/Boese0601/ByteMorph-Demo}}
\checkdata[Project Page]{\url{https://boese0601.github.io/bytemorph}}
\checkdata[Benchmark]{\url{https://huggingface.co/datasets/ByteDance-Seed/BM-Bench}}
\checkdata[Dataset]{\url{https://huggingface.co/datasets/ByteDance-Seed/BM-6M}}
\checkdata[Code]{\url{https://github.com/ByteDance-Seed/BM-code}}

\begin{document}

\setlength{\headheight}{33.04727pt}
\addtolength{\topmargin}{-21.04727pt}

\maketitle


\begin{figure}[ht]
\centering
\includegraphics[width=\linewidth]{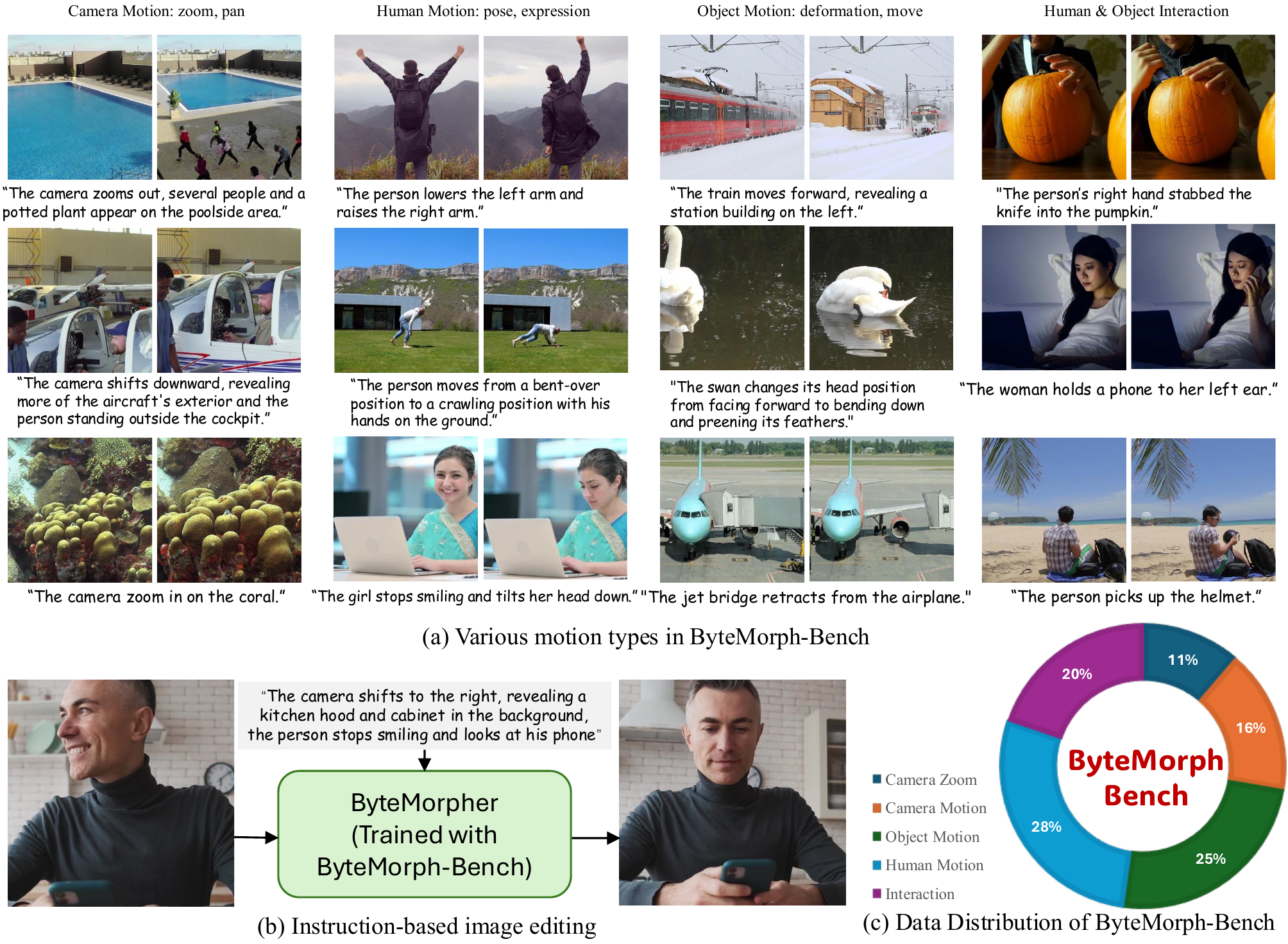}
\caption{\textbf{Overview of \framework}. a) We first construct \dataset and \benchmark, a large-scale dataset and a corresponding evaluation benchmark with data that covers a diverse range of non-rigid motions.
b) We then fine-tune \model, a diffusion transformer model initialized with pretrained weights from Flux.1-dev~\cite{flux1dev}, using the data collected in \dataset.  The distribution of \dataset is shown in c).
}
\label{fig:method}
\end{figure}

\section{Introduction}\label{sec:intro}
The rapid advancement of multi-modal datasets~\citep{lin2014microsoft,schuhmann2022laion} and generative modeling techniques~\citep{goodfellow2014generative,dinh2014nice} has substantially enhanced the capabilities of text-to-image (T2I) generation models~\citep{rombach2022high,saharia2022photorealistic,yu2023interactive,yang2024mastering}. Building upon these advancements, instruction-based image editing methods~\citep{brooks2023instructpix2pix, zhao2024ultraedit, zhang2024magicbrush, sheynin2024emu, zhang2023hive} have emerged as powerful tools that enable intuitive visual modifications through natural language instructions without necessitating explicit masks or annotations. Despite this progress, existing instruction-guided editing techniques primarily address static or localized edits, and largely overlook the complexities associated with non-rigid motion edits, such as dynamic camera movements, deformable object transformations, human articulation, and interactions between humans and objects.
While recent works have contributed datasets tailored for instruction-following editing~\citep{brooks2023instructpix2pix,zhang2024magicbrush,fu2023guiding,zhang2024magicbrush,zhao2024ultraedit,sheynin2024emu,yu2024anyedit}, they predominantly focus on appearance-centric alterations and fail to adequately capture dynamic spatial relations and broader scene transformations. Consequently, these limitations prevent models from acquiring the nuanced, motion-oriented editing capabilities necessary for realistic and expressive manipulation of visual content.

To bridge this significant gap, we propose \framework, a comprehensive framework designed explicitly for instruction-based image editing focusing on non-rigid motions. Central to \framework is \dataset, a large-scale benchmark dataset comprising diverse and high-quality image editing examples that explicitly incorporate dynamic transformations from video, including camera movements, pose adjustments, object dynamics, and scene evolutions driven by interactions. Each data instance in \dataset consists of a source image, a motion-focused natural language instruction, the corresponding edited image, and descriptive captions for both source and edited images. To build the dataset, we first generate diverse and expressive videos using an image-to-video model. We then extract motion-aware frames and pair them with automatically generated instructions that describe the intended transformations. This pipeline ensures that the generated data is both semantically rich and visually realistic.
For rigorous assessment, we additionally provide a curated evaluation benchmark, \benchmark, containing 613 challenging test samples spanning various categories of non-rigid motion edits.

Leveraging \dataset, we further introduce \model, a Diffusion Transformer baseline model specifically tailored for motion-aware editing tasks. \model integrates multi-modal attention mechanisms, effectively capturing complex non-rigid transformations guided by natural language instructions. Through extensive evaluations across multiple dynamic editing scenarios, we demonstrate that \model substantially surpasses current open-source methods~\citep{tan2024ominicontrol,cao2024instruction,zhao2024ultraedit,brooks2023instructpix2pix,zhang2024magicbrush,yu2024anyedit}, particularly excelling in edits involving viewpoint adjustments, articulated movements, and multi-object interactions. Our experiments also highlight the challenges of existing foundation models from the industry~\citep{liu2025step1x-edit,deepmind_gemini_flash,openai2025introducing4o,shi2024seededit,hidream_e1,imagen3} in preserving realism and accurately following instructions within dynamic editing contexts, thereby emphasizing the necessity for specialized datasets such as \dataset and further research for building advanced models upon \model.

Our contributions are summarized as follows:

\begin{itemize}
\item We introduce \framework, a unified framework for expressive and instruction-based image editing encompassing non-rigid motions.
\item We present \dataset and \benchmark, a large-scale dataset and a comprehensive benchmark, with high-quality image pairs for training and evaluation, addressing various dynamic editing scenarios, including camera motion, object transformation, human articulation, and human-object interaction.
\item With the proposed dataset, we finetune \model, a DiT-based model specifically developed for instruction-guided, motion-centric image editing, setting a baseline for performance on non-rigid motion editing tasks.
\end{itemize}

\begin{table*}[ht]
    \centering
    \resizebox{\textwidth}{!}{%
    \begin{tabular}{l|c|cccc|c|cc}
    \hline
    \multirow{3}{*}{\textbf{Dataset}} & \multirow{3}{*}{\textbf{Size}} & \multicolumn{4}{c|}{\textbf{Non-Rigid Edit}} & \multirow{1}{*}{\begin{tabular}{@{}c@{}}\textbf{Other Edits})\end{tabular}} & \multicolumn{2}{c}{\textbf{Image Pair Generation}} \\
    \cline{3-9}
    & & \multicolumn{1}{c|}{\scriptsize\begin{tabular}{@{}c@{}}Human\\Pose\end{tabular}} & \multicolumn{1}{c|}{\scriptsize\begin{tabular}{@{}c@{}}Object\\Deform.\end{tabular}} & \multicolumn{1}{c|}{\scriptsize\begin{tabular}{@{}c@{}}H-O\\Interact.\end{tabular}} & \multicolumn{1}{c|}{\scriptsize\begin{tabular}{@{}c@{}}Camera\\Motion\end{tabular}} & \tiny (Stylize, Attr. Mod.) & \multicolumn{1}{c}{\scriptsize\begin{tabular}{@{}c@{}}Sequential\\(From Video)\end{tabular}} & \multicolumn{1}{c}{\scriptsize\begin{tabular}{@{}c@{}}Model-Edited\\(From Images)\end{tabular}} \\ 
    \hline
    MagicBrush~\cite{zhang2024magicbrush} & 10K & \mycrossmark & \mycrossmark & \mycrossmark & \mycrossmark & \mycheckmark & \mycrossmark & \mycheckmark \\
    InstructPix2Pix~\cite{brooks2023instructpix2pix} & 313K & \mycrossmark & \mycrossmark & \mycrossmark & \mycrossmark & \mycheckmark & \mycrossmark & \mycheckmark \\
    HQ-Edit~\cite{hui2024hq} & 197K & \mycrossmark & \mycrossmark & \mycrossmark & \mycrossmark & \mycheckmark & \mycrossmark & \mycheckmark \\
    EditWorld~\cite{yang2024editworld} & 8.6K & \mycrossmark & \newcheckmark & \mycrossmark & \mycrossmark & \mycheckmark & \mycrossmark & \mycheckmark \\
    UltraEdit~\cite{zhao2024ultraedit} & 4M & \mycrossmark & \mycrossmark & \mycrossmark & \mycrossmark & \mycheckmark & \mycrossmark & \mycheckmark \\
    AnyEdit~\cite{yu2024anyedit} & 2.5M & \mycheckmark & \mycheckmark & \mycrossmark & \newcheckmark & \mycheckmark & \mycrossmark & \mycheckmark \\
    \hline \hline
    \rowcolor{gray!15}
    \dataset & \textbf{6.4M} & \mycheckmark & \mycheckmark & \mycheckmark & \mycheckmark & \mycrossmark & \mycheckmark & \mycrossmark \\
    \hline
    \end{tabular}%
    }
        \caption{Comparative analysis of popular publicly available instruction-based image editing datasets. \dataset has a dedicated \textbf{focus on non-rigid image editing}. Its image pairs are uniquely \textbf{derived sequentially from video data}, ensuring natural content consistency and realistic transformations, ideal for complex motion editing tasks. It provides extensive coverage (6.4M total images including Human Pose, Object Deformation, H-O Interactions combined, and Camera Motions, all marked with \mycheckmark), while other edit types (stylization,  attribute modification) are not its primary focus.\\ \mycheckmark: Supported. \mycrossmark: Not Supported. \newcheckmark: Partial/Minor Support.
    }
    \label{table:detailed_comparison_pair_generation}
\end{table*}
\section{Related Works}\label{sec:relatedwork}

\subsection{Image Editing with Diffusion} 
Diffusion models~\citep{sohl2015deep,song2019generative,ho2020denoising} have become a cornerstone of text-to-image synthesis, adept at progressively converting random noise into detailed visual content. Building on this foundation, various image editing techniques have emerged. These primarily operate by manipulating sampling trajectories~\citep{kim2022diffusionclip,meng2021sdedit,couairon2022diffedit,mokady2023null,kwon2022diffusion,parmar2023zero,huberman2024edit,brack2024ledits++}, modifying internal model mechanisms or feature representations~\citep{hertz2022prompt,tumanyan2023plug,cao2023masactrl}, or incorporating iterative optimization~\citep{kawar2023imagic,li2023layerdiffusion,zhang2023forgedit}.
Key strategies include sampling-based methods like SDEdit~\citep{meng2021sdedit}, which guide image reconstruction from noisy inputs via text prompts, often without retraining. Another prominent category involves attention map manipulation (e.g., Prompt-to-Prompt~\citep{hertz2022prompt}, Plug-and-Play~\citep{tumanyan2023plug}, MasaCtrl~\citep{cao2023masactrl}). These typically rely on DDIM inversion~\citep{song2020denoising}, the quality of which is crucial and has been significantly enhanced by refinement techniques such as Null-text Inversion~\citep{mokady2023null}, Direct Inversion~\citep{ju2023direct}, ProxEdit~\citep{han2024proxedit}, and TurboEdit~\citep{deutch2024turboedit,wu2024turboedit}.

\subsection{Datasets for Instruction-based Editing} 
Instruction-based image editing models, which aim to efficiently edit real images according to user guidance, rely critically on the quality and scope of their training datasets. Table.~\ref{table:detailed_comparison_pair_generation} summarizes popular public available datasets, including InstructPix2Pix~\citep{brooks2023instructpix2pix}, MagicBrush~\citep{zhang2024magicbrush}, HQ-Edit~\citep{hui2024hq}, EditWorld~\citep{yang2024editworld}, UltraEdit~\citep{zhao2024ultraedit}, and AnyEdit~\citep{yu2024anyedit}. While some datasets like EditWorld and UltraEdit incorporate human curation to improve quality, many others such as InstructPix2Pix, MagicBrush, and HQ-Edit, rely heavily on automated data-generation pipelines.

These pioneering datasets, however, are limited in the following aspects.
\textbf{First}, they largely prioritize general image editing tasks (\textit{e.g.}, stylization, attribute modification) which aim to preserve the overall image structure. This prevalent focus means they often neglect critical non-rigid transformations, such as changes in object pose, viewpoint, or dynamic camera movements. This limitation is evident in datasets like SEED-Data-Edit~\citep{ge2024seed}, UltraEdit~\citep{zhao2024ultraedit}, EMU-Edit~\citep{sheynin2024emu}, EditWorld~\citep{yang2024editworld}, and AnyEdit~\citep{yu2024anyedit}.
\textbf{Second}, the common practice for generating image pairs involves applying tuning-free editing algorithms to single source images (either real or synthetic). This approach can lead to inconsistent quality in the target images, which may contain artifacts and lack the naturalness inherent in true sequential changes. These collective shortcomings impede the development of models capable of performing robust, realistic, and complex non-rigid edits.

To overcome these deficiencies, \dataset introduces a dataset specifically tailored for instruction-based editing involving non-rigid motions. Crucially, image pairs in \dataset are derived in a way that ensures high fidelity and natural consistency, directly addressing the quality concerns of previous datasets.


\subsection{Video Generation for Image Editing} 
The integration of video data into image editing methodologies remains a relatively nascent area of research. Existing approaches that leverage video can be broadly categorized. One common strategy involves extracting frame pairs from pre-existing videos, often to capture subjects under diverse conditions or to represent changes over time~\citep{chen2024anydoor,alzayer2024magic,shi2024instadrag,luo2024readout,chen2024unireal}. A more targeted example, InstructMove~\citep{cao2024instruction}, also extracts frame pairs from real videos but further employs multi-modal Large Language Models (MLLMs) to generate corresponding editing instructions. However, this approach is often hampered by the inaccuracy of these generated instructions, stemming from the inherent difficulties MLLMs face in precisely interpreting complex transformations within dynamic video sequences.
An alternative direction reformulates image editing itself as a video generation task. For instance, Frame2Frame~\citep{rotstein2024pathways} generates a video sequence based on a source image and an editing instruction, from which a desired edited frame is then selected. While innovative, this method typically encounters significant challenges in terms of the generation quality of individual frames and overall computational efficiency.

In contrast to these approaches, our work leverages high-fidelity video synthesis as a foundational step for creating a rich data source. We employ Seaweed~\citep{seawead2025seaweed}, a state-of-the-art and highly efficient video generation model, to synthesize diverse and high-quality videos. These synthesized videos then serve as the primary data for training our baseline model. Our data significantly enhances the editing model's ability to learn and generate realistic camera movements, complex object dynamics, and intricate scene-level transformations. By addressing the data limitations and methodological challenges of prior work, our approach advances the capability of image editing systems to handle dynamic and complex scenarios more effectively.

\begin{figure}
\centering
\includegraphics[width=\linewidth]{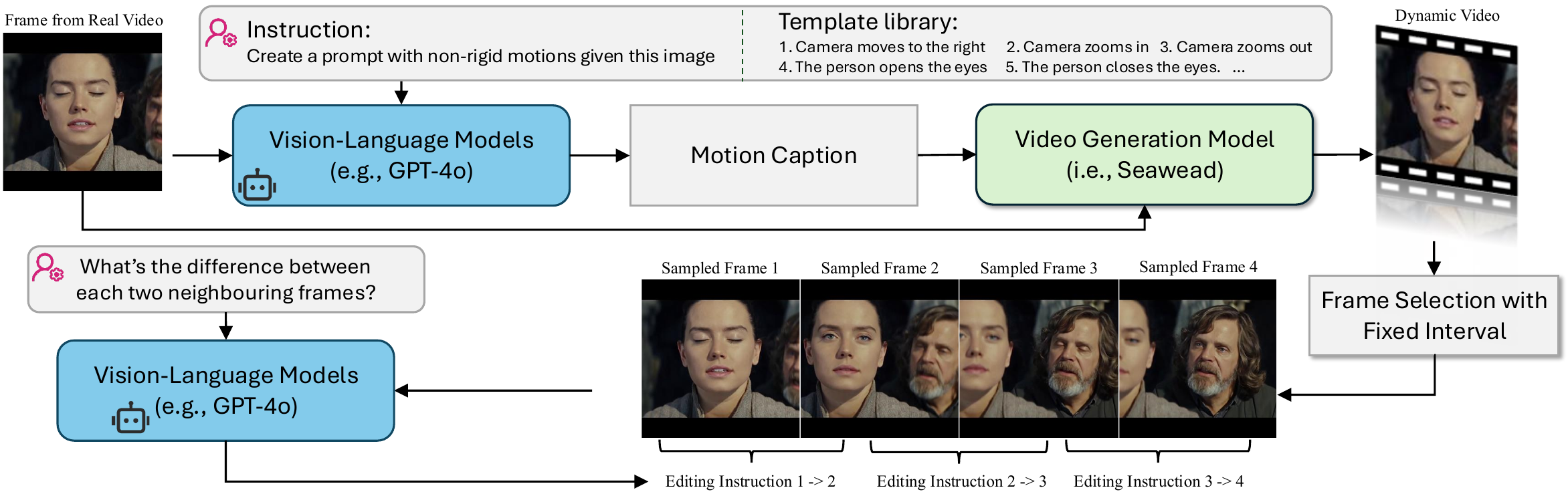}
\caption{\textbf{Overview of Synthetic Data Construction}. Given a source frame extracted from the real video, our pipeline proceeds in three steps. a) A Vision-Language Model (VLM) creates a Motion Caption from the instruction template database to animate the given frame. b) This caption guides a video generation model~\citep{seawead2025seaweed} to create a natural transformation. c) We sampled frames uniformly from the generated dynamic videos with a fixed interval and treated each pair of neighbouring frames as an image editing pair. We re-captioned the editing instruction by the same VLM, as well as the general description of each sampled frame (not shown in the figure).
}

\label{fig:method_image}
\end{figure}

\section{Dataset Construction}\label{sec:dataset}

In this section, we introduce \dataset, a novel dataset designed to model image editing as a motion transformation process. Our methodology leverages video generation models to produce natural and coherent transitions between source and target images, ensuring edits are both realistic and motion-consistent. The synthetic data construction for our dataset involves three key stages, as depicted in Figure.~\ref{fig:method_image}.

\subsection{Motion Caption} \label{sec:caption} 
Conventional text-based image editing methods typically employ a source image $I_{src}$ paired with an editing instruction prompt $P$, which specifies the desired modifications. In contrast, we treat editing as a sequential transformation, introducing a prompt type—Motion Caption ($C_{m}$)—that explicitly describes the progression of edits over time.

Given a source frame $I_{src}$ extracted from a real video $V_{src}$, we integrate visual features from $I_{src}$ with the editing instruction prompt $P$ from a template prompt library, to formulate $C_{m}$ which is a narrative description that captures the dynamic evolution of the intended edits. The template prompt library contains diverse motion-aware editing prompts for humans, objects, and camera viewpoints and this strategy was inspired by in-context learning (ICL) techniques~\citep{dong2022survey}, as introduced by~\citep{rotstein2024pathways}. This caption generation process is automated by leveraging a Vision-Language Model (VLM), particularly ChatGPT-4o~\citep{openai2024chatgpt4o} in our case. The VLM generates concise, video-like narratives describing transformations.

\begin{table*}[!t]
\centering
\resizebox{0.7\linewidth}{!}{
    \begin{tabular}{l l | c c | c c | c}
    \hline
    && CLIP-SIM$_{txt}$$\uparrow$ & \cellcolor[HTML]{FFCC99}{CLIP-D$_{txt}$$\uparrow$} & CLIP-SIM$_{img}$$\uparrow$ & \cellcolor[HTML]{FFB366}{CLIP-D$_{img}$$\uparrow$}& \cellcolor[HTML]{FFB366}{VLM-Eval$\uparrow$}  \\
    \hline
        \multirow{9}{*}{\rotatebox[]{90}{\small{Camera Zoom}}}         &   InstructPix2Pix~\citep{brooks2023instructpix2pix}& 0.270 & 0.021 & 0.737 &0.266 & 42.37 \\
        &MagicBrush~\citep{zhang2024magicbrush} & \textbf{0.311} & 0.002 & \underline{0.907}& 0.202 & 49.37 \\
        &UltraEdit (SD3)~\citep{zhao2024ultraedit}  & 0.299& 0.000 & 0.864&0.249 & 54.74 \\
        & AnySD~\cite{yu2024anyedit} & 0.309 &0.001 & \textbf{0.911} &0.182 & 40.92  \\
        & InstrcutMove~\cite{cao2024instruction} &0.283 & 0.027 & 0.821 & 0.294 & 70.66 \\

        & OminiControl~\cite{tan2024ominicontrol} & 0.251 & 0.022& 0.722 & 0.300& 45.79 \\
        &  $\dagger$InstrcutMove~\cite{cao2024instruction} & 0.301 & \underline{0.045} & 0.846 & \underline{0.425} & \underline{82.29} \\
        & $\dagger$OminiControl~\cite{tan2024ominicontrol}  &\underline{0.310} &0.039 & 0.801& 0.414& 74.15  \\
        &$\dagger$\textbf{\model} (Ours)  & 0.301 & \textbf{0.048} & 0.847 & \textbf{0.463} &  \textbf{84.08} \\
        \hline
         & GT & 0.317 &  0.075 & 0.890 & 1.000 &  87.11 \\
    \hline
    \hline
        \multirow{9}{*}{\rotatebox[]{90}{\small{Camera Move}}}       & InstructPix2Pix~\citep{brooks2023instructpix2pix} & \underline{0.318} & 0.010& 0.709& 0.200 & 32.20 \\
        &MagicBrush~\citep{zhang2024magicbrush}   & 0.317& 0.009& \textbf{0.913} & 0.195 & 52.63   \\
        &UltraEdit (SD3)~\citep{zhao2024ultraedit} & 0.306 & 0.012 & 0.885 & 0.240 & 59.01 \\
        & AnySD~\cite{yu2024anyedit} & 0.318 & 0.010& \underline{0.909} & 0.200 & 49.37 \\
        & InstrcutMove~\cite{cao2024instruction} & 0.305 &0.016&0.862  &0.291 & 74.86 \\
        
        & OminiControl~\cite{tan2024ominicontrol} & 0.243 & 0.022 & 0.687 & 0.243 & 16.71 \\
        &  $\dagger$InstrcutMove~\cite{cao2024instruction} & 0.304 &\underline{0.027}& 0.883 &\underline{0.412} & \underline{82.53} \\
        & $\dagger$OminiControl~\cite{tan2024ominicontrol}  & 0.298& 0.025& 0.891 & 0.304 & 79.26 \\
        &$\dagger$\textbf{\model} (Ours)  & \textbf{0.319} & \textbf{0.041}& 0.894& \textbf{0.426}& \textbf{84.18} \\
        \hline
         & GT & 0.320 &0.039 & 0.915 & 1.000 & 86.37 \\
        \hline
    \hline
        \multirow{9}{*}{\rotatebox[]{90}{\small{Object Motion}}} &      InstructPix2Pix~\citep{brooks2023instructpix2pix}& 0.299& 0.026& 0.789 & 0.257 & 36.47\\
        &MagicBrush~\citep{zhang2024magicbrush} & 0.328 &0.007 & \textbf{0.901} &0.163 &  47.49 \\
        &UltraEdit (SD3)~\citep{zhao2024ultraedit}  & 0.324 & 0.012& 0.887& 0.237& 62.13 \\
        & AnySD~\cite{yu2024anyedit} & 0.319 & 0.008 & 0.879 & 0.189 & 48.31 \\
        & InstrcutMove~\cite{cao2024instruction} &0.325 &0.015 &0.870  &0.318 & 72.44 \\
        
        & OminiControl~\cite{tan2024ominicontrol} & 0.279 & 0.023 & 0.753 &0.270 & 34.11\\
        &  $\dagger$InstrcutMove~\cite{cao2024instruction} & 0.328 & \underline{0.043} & 0.891 & \textbf{0.481} &\underline{87.97} \\
        & $\dagger$OminiControl~\cite{tan2024ominicontrol}  &\underline{0.330} & 0.036& 0.892 &0.470 & 86.48 \\
        &$\dagger$\textbf{\model} (Ours)  &  \textbf{0.332} & \textbf{0.044} & \underline{0.896} & \underline{0.472} & \textbf{89.07} \\
                \hline
         & GT & 0.335 &0.056 & 0.919 & 1.000 & 89.53 \\
        \hline
    \hline
        \multirow{9}{*}{\rotatebox[]{90}{\small{Human Motion}}} 
        &
      InstructPix2Pix~\citep{brooks2023instructpix2pix} &0.248 & 0.012& 0.694 & 0.211& 23.60  \\
        &MagicBrush~\citep{zhang2024magicbrush}   & \textbf{0.317} & 0.001 & \textbf{0.911} & 0.146 & 46.27 \\
        &UltraEdit (SD3)~\citep{zhao2024ultraedit}  & 0.313 & 0.011& 0.900 & 0.195 & 50.64 \\
        & AnySD~\cite{yu2024anyedit} &  0.312 &0.003 &0.894 & 0.156&38.12 \\
        & InstrcutMove~\cite{cao2024instruction} & 0.308 &  0.013 & 0.861 & 0.278& 69.43\\
        
        & OminiControl~\cite{tan2024ominicontrol} & 0.230 & 0.018 & 0.660 & 0.229 & 25.18  \\
        &  $\dagger$InstrcutMove~\cite{cao2024instruction} & 0.314 & \textbf{0.023} & \underline{0.901} & \textbf{0.442} & \underline{84.70} \\
        & $\dagger$OminiControl~\cite{tan2024ominicontrol}  & 0.311 & 0.016& 0.880 &0.399& 80.78 \\
        &$\dagger$\textbf{\model} (Ours)  &  \underline{0.316} & \underline{0.022}& 0.899 & \underline{0.440}& \textbf{85.66} \\
                \hline
        & GT & 0.321 & 0.031 &0.922 &1.000 & 86.10 \\
        \hline
    \hline
        \multirow{9}{*}{\rotatebox[]{90}{\small{Interaction}}} 
        &
        InstructPix2Pix~\citep{brooks2023instructpix2pix} &0.271 & 0.020& 0.732 & 0.263 &31.29 \\
        &MagicBrush~\citep{zhang2024magicbrush}   & \underline{0.317} & 0.004 & \textbf{0.914} & 0.167& 39.98 \\
        &UltraEdit (SD3)~\citep{zhao2024ultraedit}  & 0.314 & 0.018& 0.892 &0.226 & 52.24 \\
        & AnySD~\cite{yu2024anyedit} & 0.315& 0.005 & \underline{0.909} &0.173 & 37.23\\
        & InstrcutMove~\cite{cao2024instruction} &0.309 &0.019 &0.855  &0.318 &67.07 \\
        
        & OminiControl~\cite{tan2024ominicontrol} & 0.258 & 0.021 & 0.689 & 0.265 & 32.99 \\
        & $\dagger$InstrcutMove~\cite{cao2024instruction} & 0.314& \underline{0.043} & 0.885 & \underline{0.477}& \underline{85.83} \\
        & $\dagger$OminiControl~\cite{tan2024ominicontrol}  & 0.295 & 0.041 & 0.768 & 0.433 & 78.90 \\
        &$\dagger$\textbf{\model} (Ours)  &  \textbf{0.320} & \textbf{0.045} &0.884 & \textbf{0.483} & \textbf{86.61} \\
                \hline
        & GT & 0.324 & 0.046 & 0.905 & 1.000 & 88.84  \\
        \hline
    \hline

    \end{tabular}
}
\caption{Quantitative Evaluation of open-source methods on \benchmark. $\dagger$ represents that the method is trained on \dataset. We run each method four times and report the average value. In addition to CLIP similarity metrics, we use Claude-3.7-Sonnet~\cite{claude37sonnet} to evaluate the overall editing quality (VLM-Score) on a scale of 0-100. The metrics highlighted in \colorbox[HTML]{FFCC99}{light orange} and \colorbox[HTML]{FFB366}{dark orange} are more important.
}\label{table:2}
\end{table*}

\subsection{Video Generation} \label{sec:video_generation}

We utilize Seaweed~\citep{seawead2025seaweed}, a pretrained generative video latent diffusion model built upon a transformer architecture, to simulate dynamic editing processes. Seaweed is fine-tuned to generate a coherent video sequence conditioned on both the source image $I_{src}$ and a corresponding image-to-video prompt, which is the motion caption $C_{m}$ in this case.
Initially, $I_{src}$ is encoded into a latent representation. Subsequently, guided by $C_{m}$, the model performs a denoising operation, synthesizing a temporally coherent frame sequence that embodies the desired editing transformation. The transformer backbone effectively integrates visual and textual inputs, ensuring accurate control and consistency throughout the editing process.
Given the video generation model $G$, the video generation can be expressed as: $G(I_{src}, C_{m}) = V_{gen} = \{F_1, \dots, F_T\}$, where $V_{gen}$ represents the generated video sequence consisting of $T$ frames, and $F_t$ denotes the frame at timestep $t$.

\subsection{Image-Instruction Pair Creation} 
\label{sec:image_instruction_pair_creation}

The generated video sequences from Seaweed~\citep{seawead2025seaweed} provide a dynamic basis for constructing high-quality image editing pairs. To effectively leverage these videos for training, we uniformly sample four frames $\widetilde{F_{i}}$ from each synthetic video sequence of length $T=301$ with fixed intervals of 100 frames, ensuring comprehensive coverage of the editing transformation dynamics.
Each pair of sampled neighboring frames $(\widetilde{F}_i, \widetilde{F}_{i+1})$ constitutes a source-target editing pair, thereby forming three distinct editing examples per generated sequence: $(\widetilde{F}_1, \widetilde{F}_2)$, $(\widetilde{F}_2, \widetilde{F}_3)$, and $(\widetilde{F}_3, \widetilde{F}_4)$. These pairs capture incremental but perceptually significant transformations, essential for modeling nuanced non-rigid motions.
To avoid the cases where generated videos did not follow the transformation described in the motion caption $C_{m}$, we utilize the same Vision-Language Model (VLM), ChatGPT-4o~\citep{openai2024chatgpt4o}, to automatically generate precise editing instructions $\widetilde{P}_{i,i+1}$. The VLM examines each frame pair and produces descriptive captions detailing the transformation from the source frame to the target frame. Additionally, a general description of each sampled frame is provided, enriching the dataset with contextual semantic information. This automated captioning ensures consistency, scalability, and semantic clarity across the dataset. 
Through this method, we effectively construct an extensive, high-quality collection of image-instruction pairs, specifically curated for benchmarking and developing advanced models capable of intricate, motion-guided image editing tasks.

\subsection{Dataset Analysis}

Our dataset contains a total of 6.45 million instruction-based image editing data samples with five motion categories, as shown in Figure.~\ref{fig:method}. To our knowledge, this is the largest dataset focused on non-rigid image editing to be released to the public (a detailed comparison to other datasets can be found in Table 1).
While existing instruction-guided image editing datasets~\citep{hui2024hq, zhang2024magicbrush, zhao2024ultraedit, sheynin2024emu} primarily emphasize appearance alterations and localized modifications, they rarely contain non-rigid motion transformations. In contrast, \dataset is explicitly designed to address this limitation by focusing on dynamic edits such as camera movements, human articulation, and object-object interactions—motions that are underrepresented in current benchmarks. To ensure realism and motion fidelity, \dataset comprises high-quality image pairs derived from realistic motion transformations. Furthermore, \dataset offers consecutive frames sampled from videos, enabling research into other tasks like multi-reference editing (detailed in Section.~\ref{sec:future}), expanding beyond the single-reference settings explored in works such as InstructMove~\citep{cao2024instruction} and Frame2Frame~\citep{rotstein2024pathways}. Importantly, improving image-based motion editing is not only scientifically valuable but also practically advantageous: image edits execute in seconds, while video generation may take minutes. This advantage of efficiency makes it possible for rapid, flexible customization in real-world applications where user responsiveness and iteration are critical.

\subsection{Model Finetuning}\label{sec:method}

\begin{figure}[!ht]
\centering
\includegraphics[width=\linewidth]{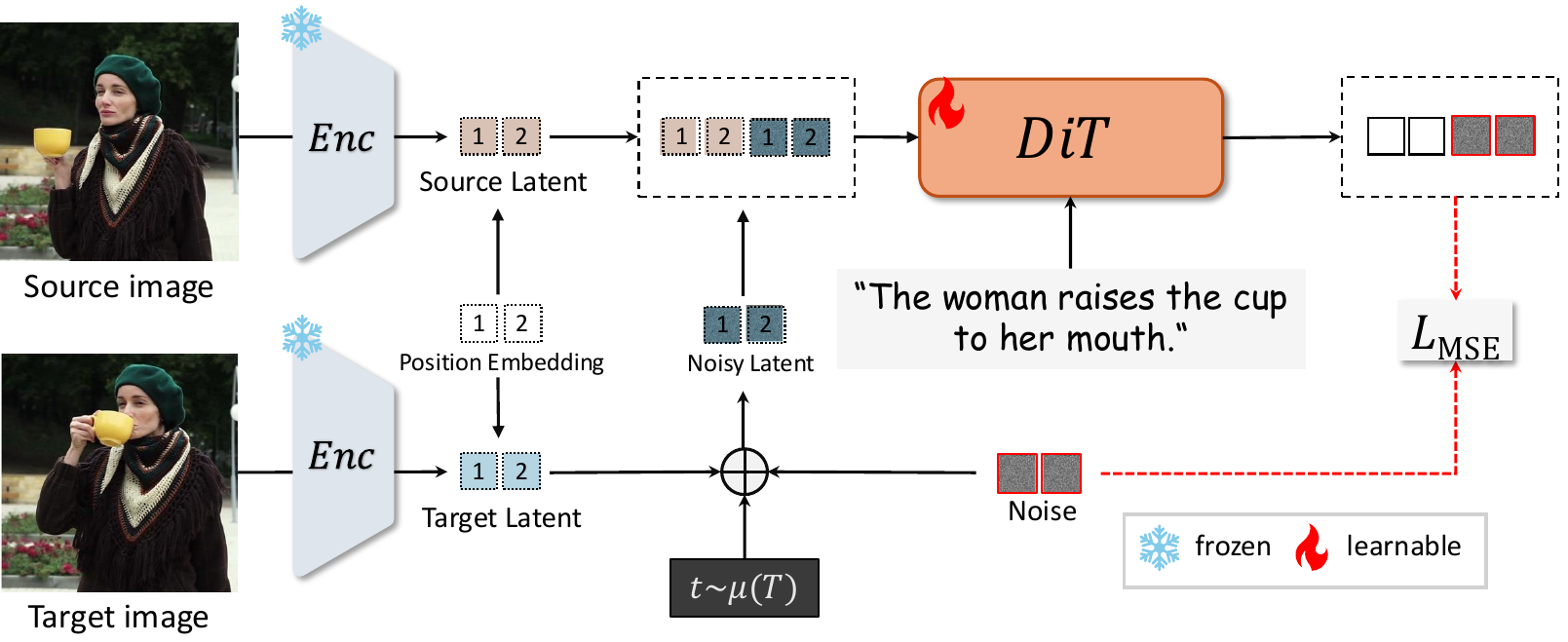}

\caption{\textbf{Overview of \model's training.} We fine-tuned the Diffusion Transformer backbone from the pre-trained Flux.1-dev~\citep{flux1dev} text-to-image model on the \dataset. We feed the source image and target image into the same frozen VAE encoder and obtain source and target latents. The position encoding for both latents is shared with exactly the same embeddings. The target latent is added with noise and further concatenated with the source latent along the sequence length dimension. The DiT is fine-tuned with the MSE loss, where the difference between the noise and the latter half of the output from DiT (corresponding to the noisy target latent input) is minimized. 
}

\label{fig:bytemorpher}
\end{figure}

The overall fine-tuning pipeline of \model is shown in Figure~\ref {fig:bytemorpher}.  After construction of \dataset, we fine-tuned a Diffusion Transformer Model on the training data with the pre-trained Flux.1-dev~\citep{flux1dev} text-to-image model as the backbone. We feed the source image and target image into the same VAE encoder and concatenate the noisy target latent and source latent along the sequence length dimension. The position encoding for both latents is shared with exactly the same embeddings. The DiT is fine-tuned with the same loss as the original Flux.1-dev, while all other parameters are frozen.

\begin{table}[!t]
\centering
\resizebox{0.95\textwidth}{!}{
    \begin{tabular}{l l | c c | c c | c c c }
    \hline
    && CLIP-SIM$_{txt}$$\uparrow$ & \cellcolor[HTML]{FFCC99}{CLIP-D$_{txt}$$\uparrow$} & CLIP-SIM$_{img}$$\uparrow$ & \cellcolor[HTML]{FFB366}{CLIP-D$_{img}$$\uparrow$}& \cellcolor[HTML]{FFB366}{VLM-Eval$\uparrow$}  & \cellcolor[HTML]{FFB366}{Human-Eval-FL$\uparrow$} & \cellcolor[HTML]{FFB366}{Human-Eval-ID$\uparrow$} \\
    \hline
        \multirow{11}{*}{\rotatebox[]{90}{\small{Camera Zoom}}}         
         
        & Step1X-Edit~\cite{liu2025step1x-edit} & 0.310 & 0.025 &\textbf{0.943}  & 0.258 & 59.34 & 26.60 & 48.86  \\
        & HiDream-E1-FULL~\cite{hidream_e1} & 0.304 & 0.027 & 0.682 & 0.287 & 41.18 &  33.00 & 16.50 \\
        & Imagen-3-capability~\cite{imagen3} & 0.293 & 0.025 &0.846  & 0.264 & 53.94 &  \underline{61.34} &41.38  \\   
        & Gemini-2.0-flash-image~\cite{deepmind_gemini_flash} & 0.305 & {0.031}& {0.862} & 0.297 & 72.27 & 61.04 &63.09  \\
        & SeedEdit 1.6~\cite{shi2024seededit} & {0.311} & 0.029 &  0.827 & 0.325 & 75.00 & \underline{61.34}& \textbf{83.60} \\
        & GPT-4o-image~\cite{openai2025introducing4o} & \textbf{0.317} & 0.015 & 0.832 & {0.337} & \underline{88.14} &  \textbf{89.36} & 61.09  \\
        & BAGEL~\cite{deng2025bagel} & 0.300 &0.031 &0.860 & 0.301 & 75.55 & - & -  \\
        & Flux-Kontext-pro~\cite{fluxkontext2025}& \underline{0.312} &0.024 & 0.864 &0.334 & 75.66 & - & -  \\
        & Flux-Kontext-max~\cite{fluxkontext2025}& 0.307 & \underline{0.032} &\underline{0.871} &\underline{0.373} & 80.18 & - & -  \\
        & SeedEdit 3.0~\cite{wang2025seededit} & 0.296&0.027 &0.833 &0.370 & \textbf{88.25} & -&- \\
        &\textbf{\model} (Ours)  & 0.301 & \textbf{0.048} & 0.847 & \textbf{0.463} &  {84.08} &61.13	 & \underline{74.73} \\

        \hline
        & GT & 0.317 &  0.075 & 0.890 & 1.000 &  87.11 & - & - \\
        \hline
        \hline
        \multirow{11}{*}{\rotatebox[]{90}{\small{Camera Move}}}    
         
        & Step1X-Edit~\cite{liu2025step1x-edit} & 0.315 & 0.008 &  \textbf{0.946} &0.208 & 57.96 & 33.50 & 63.39 \\
        & HiDream-E1-FULL~\cite{hidream_e1} & 0.309 & \underline{0.029}  &0.712 &0.252 & 32.76 & 16.50 & 18.22\\
        & Imagen-3-capability~\cite{imagen3} & 0.282& 0.010 & 0.813 &0.238 & 47.22& 17.38 & 26.51  \\
        & Gemini-2.0-flash-image~\cite{deepmind_gemini_flash} & 0.317 & 0.020  & 0.892& {0.311} &77.96 & 56.60& \underline{75.76} \\
        & SeedEdit 1.6~\cite{shi2024seededit} & 0.314 & 0.015  &0.866 & 0.253 &  78.59&58.30 & \textbf{87.78}\\
        & GPT-4o-image~\cite{openai2025introducing4o} & \textbf{0.321} & 0.011 & 0.865 & 0.285 & \underline{84.57}& \textbf{76.74} & 59.14 \\
                & BAGEL~\cite{deng2025bagel} &0.306 &0.026 &0.883 & 0.290&76.08 &- &- \\
        & Flux-Kontext-pro~\cite{fluxkontext2025}  & 0.312&0.016 &0.891 &0.286 &79.14 & -&-  \\
        & Flux-Kontext-max~\cite{fluxkontext2025}  & 0.315&0.019 & \underline{0.896} & \underline{0.325} & \textbf{85.97}& -&-  \\
        & SeedEdit 3.0~\cite{wang2025seededit} & 0.308 & 0.020& 0.887& 0.278& 78.00& -&- \\
          &\textbf{\model} (Ours)  & \underline{0.319} & \textbf{0.041}& {0.894}& \textbf{0.426}& {84.18} & \underline{67.60}	& 58.25\\

        \hline
       & GT & 0.320 &0.039 & 0.915 & 1.000 & 86.37 &  - & - \\
    \hline
    \hline
        \multirow{11}{*}{\rotatebox[]{90}{\small{Object Motion}}} 
         
        & Step1X-Edit~\cite{liu2025step1x-edit} & 0.323 &0.019 & \textbf{0.923} &0.260 & 72.78 &72.16 & 59.39  \\
        & HiDream-E1-FULL~\cite{hidream_e1} & 0.312& 0.028 & 0.700& 0.259& 35.00 & 44.34 & 49.75 \\
        & Imagen-3-capability~\cite{imagen3} &0.324 &0.027  & 0.870 &0.261 & 57.06 & 62.56&77.84 \\
        & Gemini-2.0-flash-image~\cite{deepmind_gemini_flash} & \underline{0.333} & \underline{0.040} &0.892  & {0.341}  & 79.08 & \underline{74.77} & \textbf{86.62} \\ 
        & SeedEdit 1.6~\cite{shi2024seededit} &0.332 &0.025  & 0.874 &0.323 & 80.21 &66.50 & \underline{79.12}  \\
        & GPT-4o-image~\cite{openai2025introducing4o} & \textbf{0.339}& 0.029  &0.861  &\underline{0.354} & \textbf{90.60} & \textbf{75.19} & 49.91 \\
        & BAGEL~\cite{deng2025bagel} &0.324&0.036 &\underline{0.920} &0.326 &74.07 &- &-   \\
        & Flux-Kontext-pro~\cite{fluxkontext2025}  & 0.321&0.018 &0.893 &0.314 &78.41 & -&- \\
        & Flux-Kontext-max~\cite{fluxkontext2025}  & 0.325& 0.025&0.888 & 0.353&80.42 & -&-  \\
        & SeedEdit 3.0~\cite{wang2025seededit} &0.321 &0.036 &0.905 &0.344 &88.11 & -&- \\
         &\textbf{\model} (Ours)  &  0.332 & \textbf{0.044} & {0.896} & \textbf{0.472} & \underline{89.07} & 62.16	& 58.25\\
        \hline
        & GT & 0.335 &0.056 & 0.919 & 1.000 & 89.53 &  - & -\\
    \hline
    \hline
        \multirow{11}{*}{\rotatebox[]{90}{\small{Human Motion}}} 
         
        & Step1X-Edit~\cite{liu2025step1x-edit} & 0.315 & 0.017 & \textbf{0.931}  &0.212 &  65.39 &44.50 &78.80  \\
        & HiDream-E1-FULL~\cite{hidream_e1} & 0.301 & 0.017 &0.676 &0.215 & 33.21 & 12.51 & 38.66 \\        
        & Imagen-3-capability~\cite{imagen3} & 0.295& 0.017 & 0.840 & 0.233 & 55.70& 33.34&61.17\\
        & Gemini-2.0-flash-image~\cite{deepmind_gemini_flash} &  0.314& 0.017&0.893 & {0.282} & 78.72 & 51.84& 63.34  \\
        
        & SeedEdit 1.6~\cite{shi2024seededit} & \textbf{0.324} & \underline{0.024} & 0.878 & 0.274 & 80.62 &56.23 & \underline{72.12}   \\
        & GPT-4o-image~\cite{openai2025introducing4o} & \underline{0.316} & {0.021} &0.850 & {0.330} &\underline{87.93} & \textbf{87.56}&57.84   \\
        & BAGEL~\cite{deng2025bagel} &0.312 &0.021 &\underline{0.929} &0.242 &74.36 & -& -\\
        & Flux-Kontext-pro~\cite{fluxkontext2025}  & 0.314&0.017 &0.918 &0.283 &79.15 & -&- \\
        & Flux-Kontext-max~\cite{fluxkontext2025}  & \underline{0.316}& 0.016&0.908 &0.307 &80.78 & -&-  \\
        & SeedEdit 3.0~\cite{wang2025seededit} & 0.313 &\textbf{0.025} &0.903 & \underline{0.343} &\textbf{88.13} & -&- \\
        &\textbf{\model} (Ours)  &  \underline{0.316} & {0.022}& {0.899} & \textbf{0.440}& {85.66} & \underline{68.38}	&	\textbf{75.00}\\
        \hline
        & GT & 0.321 & 0.031 &0.922 &1.000 & 86.10 & - & - \\
    \hline
    \hline
        \multirow{11}{*}{\rotatebox[]{90}{\small{Interaction}}} 
         
        & Step1X-Edit~\cite{liu2025step1x-edit}& 0.312 & 0.020 & \textbf{0.937} & 0.245 & 65.99 & 36.09 & 64.56\\
        & HiDream-E1-FULL~\cite{hidream_e1} & 0.307& 0.019 & 0.679 & 0.251 & 35.73&  10.60 & 38.66  \\
    & Imagen-3-capability~\cite{imagen3} & 0.307 & 0.023  & 0.863 & 0.254 & 54.78 & 47.16 & 61.59  \\
        & Gemini-2.0-flash-image~\cite{deepmind_gemini_flash} &0.316 &0.027 &{0.889} & {0.327} & 76.86 & 60.70 & \underline{77.94} \\
        & SeedEdit 1.6~\cite{shi2024seededit} & \textbf{0.326} & {0.032} & 0.878 & 0.316 & 78.27 &49.78 &\textbf{80.10} \\
        & GPT-4o-image~\cite{openai2025introducing4o} & {0.318} & {0.031} & 0.851 & {0.351} & \textbf{88.65} & \textbf{81.17} & 73.72 \\
        & BAGEL~\cite{deng2025bagel} & 0.312& \underline{0.037}&\underline{0.913} &0.301 &73.16 & -&- \\
        & Flux-Kontext-pro~\cite{fluxkontext2025}  & 0.313&0.028 &0.898 &0.318 &78.58 & -&-  \\
        & Flux-Kontext-max~\cite{fluxkontext2025}  & \underline{0.320} &0.032 &0.894 &0.335 &80.12 & -&-  \\
        & SeedEdit 3.0~\cite{wang2025seededit} &0.312 &0.036 &0.894 &\underline{0.371} & 86.07& -&- \\
        & \textbf{\model} (Ours)  &  \underline{0.320} & \textbf{0.045} &0.884 & \textbf{0.483} & \underline{86.61}& \underline{69.15}	& 64.73\\

        \hline
        & GT & 0.324 & 0.046 & 0.905 & 1.000 & 88.84  & - & - \\
    \hline
    \end{tabular}
}
\caption{Quantitative Evaluation of \model and other methods from the industry on \benchmark. We run each method four times and report the average value. In addition to CLIP similarity metrics, we use Claude-3.7-Sonnet~\cite{claude37sonnet} to evaluate the overall editing quality (VLM-Score). We also ask human participants to evaluate the instruction-following quality (Human-Eval-FL) and identity-preserving quality (Human-Eval-ID). The final scores from VLM and Human are post-processed and presented on a scale of 0-100 for better understanding. The metrics highlighted in \colorbox[HTML]{FFCC99}{light orange} and \colorbox[HTML]{FFB366}{dark orange} are more important.}
\vspace{-5pt}
\label{table:3}
\end{table}

\section{Experiments}\label{sec:exp}

\subsection{Implementation Details}
\noindent\textbf{Settings.} \label{sec:settings}
We employ Seaweed~\citep{seawead2025seaweed} as the video generation backbone to synthesize image-instruction pairs for \dataset. The real videos for extracting source frames were collected from online resources, carefully filtered through video quality assessments and content safety checks. For \model, we utilize FLUX.1-dev~\citep{flux1dev} as the foundational architecture,  and train the Diffusion Transformer on 8 NVIDIA H100 GPUs with a total batch size of 8. The training process employs the AdamW optimizer with a learning rate of $1 \times 10^{-5}$. Additionally, we use DeepSpeed Stage 2 to enhance training stability and optimize memory utilization.

\noindent\textbf{Benchmark Evaluation.} We manually selected 613 high-quality editing pairs, creating the \benchmark for a more challenging and comprehensive evaluation. We categorize image editing pairs into five different classes according to the instructions, including 1) camera zoom - the camera position to take the source image zooms in or out, 2) camera move - the camera position to take the source image moves horizontally or vertically, 3) object motion - the object in the source image moves, 4) human motion - the human in the source image moves, and 5) interaction - the objects or humans in the source image interact with each other. \textbf{Notably, these pairs are not visible during training.} We also conducted an evaluation on the test set of InstrctMove~\citep{cao2024instruction} benchmark with human and object non-rigid motions. Both benchmarks evaluate editing models by comparing non-rigid motion edited results with the ground truth.

\noindent\textbf{Baselines.} For evaluation of existing methods, we adopted the following baselines: 1) Open-source state-of-the-art methods from academia, including InstructPix2Pix~\citep{brooks2023instructpix2pix}, MagicBrush~\citep{zhang2024magicbrush}, 
UltraEdit~\citep{zhao2024ultraedit},
AnySD~\citep{yu2024anyedit},
OminiControl~\citep{tan2024ominicontrol}, and InstructMove~\citep{cao2024instruction} 2) Other most recent methods from industry, including Step1X-Edit~\citep{liu2025step1x-edit}, HiDream-E1-Full~\citep{hidream_e1}, Imagen-3-capabilities~\citep{imagen3}, Gemini-2.0-flash-image-generation~\citep{deepmind_gemini_flash}, SeedEdit 1.6~\citep{shi2024seededit}, GPT-4o-image-generation~\citep{openai2025introducing4o}, BAGEL~\cite{deng2025bagel}, Flux-Kontext-pro~\cite{fluxkontext2025}, Flux-Kontext-max~\cite{fluxkontext2025} and SeedEdit 3.0~\cite{wang2025seededit}. For those models without publicly available code and weights, we directly call the model APIs following the official instructions.

\noindent\textbf{Metrics.} 
Following prior works~\citep{zhang2024magicbrush, yu2024anyedit, cao2024instruction}, we adopt CLIP similarity metrics, e.g., CLIP-SIM$_{img}$, CLIP-SIM$_{txt}$, and CLIP-D$_{txt}$, in our benchmark. Specifically, we denote the source image, ground truth target image, and generated image as $I_{src}$, $I_{tgt}$, and $I_{gen}$, and the text captions of the source and target images as $T_{src}$ and $T_{tgt}$. The details of CLIP-SIM$_{img}$, CLIP-SIM$_{txt}$, and CLIP-D$_{txt}$ are defined in the appendix.

However, we observe that these metrics do not always reliably reflect editing quality, particularly for cases involving camera motion. To address this, we introduce a novel yet simple metric, CLIP-D$_{img}$,  defined as follow equation with CLIP encoder $\mathbf{C}(\cdot)$: 
\begin{align}
\text{CLIP-D}_{\text{img}} &= \cos\left( \mathbf{C}(I_{\text{gen}}) - \mathbf{C}(I_{\text{src}}),\; \mathbf{C}(I_{\text{tgt}}) - \mathbf{C}(I_{\text{src}}) \right) \tag{Eq. 1}
\end{align}

Given that the image editing pairs in our benchmark are manually curated and quality-checked, this metric offers a more accurate measure of overall editing quality. For evaluation on the InstructMove~\citep{cao2024instruction} benchmark, we directly use the original test set and metrics.
To ensure the validity of our experimental results, we evaluate all methods by running the generation process \textbf{four} times on the benchmarks and calculating the average value of the resulting metrics.

\noindent\textbf{VLM Evaluation and Human Preference Study.} For VLM evaluation, we used the Claude-3.7-Sonnet~\citep{claude37sonnet}, instead of the GPT~\citep{openai2024chatgpt4o}, or Gemini~\citep{team2023gemini} series for fair comparison and to avoid possible leakage issues. For human evaluation, we asked 40 participants to judge the Instruction-Following quality (Human-Eval) and ID-Preserving quality (Human-Eval-ID). The details are specified in Section.~\ref {sec:user_study} of the appendix.

\subsection{Evaluation}\label{sec:eval}

\noindent\textbf{Quantitative Results.}
\label{sec:quantitative}
We conduct comprehensive evaluations of both open-source methods (Table~\ref{table:2}) and industrial-grade models (Table~\ref{table:3}) on \benchmark. For interpretability, both VLM and human evaluation scores are normalized to a 0–100 scale.
\model achieves state-of-the-art performance among open-source methods, demonstrating strong instruction-following capabilities and consistent identity preservation across all editing categories. Compared to commercial models, it also delivers competitive overall performance and exhibits a more favorable balance between instruction fidelity and identity retention. These results underscore the effectiveness of our proposed \dataset and highlight its value as a rigorous benchmark for evaluating non-rigid motion editing.
Among industrial methods, GPT-4o-image~\citep{openai2025introducing4o} excels in instruction-following, as reflected by its high Human-Eval-FL scores. However, it underperforms in identity preservation compared to SeedEdit-1.6~\citep{shi2024seededit} and Gemini-2.0-flash-image~\citep{deepmind_gemini_flash}, both of which maintain stronger visual consistency with the source content. HiDream-E1-Full~\citep{hidream_e1} and Step1X-Edit~\citep{liu2025step1x-edit}, which are partially open-sourced with inference-only access, show relatively weak performance. We hypothesize that this is due to a lack of training data with diverse and fine-grained motion transformations, such as those found in \dataset. 
\begin{table*}[!t]

\centering
\resizebox{0.48\textwidth}{!}{
\begin{tabular}{l | c c c }
\hline
& \cellcolor[HTML]{FFCC99}{CLIP-D$_{txt}$$\uparrow$} & CLIP-SIM$_{txt}$$\uparrow$ & CLIP-SIM$_{img}$$\uparrow$  \\
\hline
NullTextInversion~\cite{mokady2023null} & 0.0660 & 0.7648 & 0.9063 \\
MasaCtrl~\cite{cao2023masactrl} & 0.0436 & 0.8527 & 0.9160 \\
InstructPix2Pix~\cite{brooks2023instructpix2pix} & 0.0887 & 0.8569 & \textbf{0.9380} \\
UltraEdit~\cite{zhao2024ultraedit} & 0.0824 & 0.8571 & 0.9184 \\
MagicBrush~\cite{zhang2024magicbrush} & 0.0972 & 0.8648 & \underline{0.9318} \\
InstrcutMove~\cite{cao2024instruction} & \textbf{0.1361} & \underline{0.8724} & 0.9275 \\
OminiControl~\cite{tan2024ominicontrol} & 0.1089 & 0.8359 & 0.7909 \\
AnySD~\cite{yu2024anyedit} & 0.0463 & 0.8415 & 0.9240 \\
\hline
\textbf{\model} (Ours) & \underline{0.1335} & \textbf{0.8784} & 0.9083 \\
\hline
\end{tabular}
}
\caption{Quantitative comparison with state-of-the-art text-guided image editing methods on InstructMove non-rigid image editing benchmark.}
\label{table:1}

\end{table*}

\begin{figure}[ht]
\centering
\includegraphics[width=\linewidth]{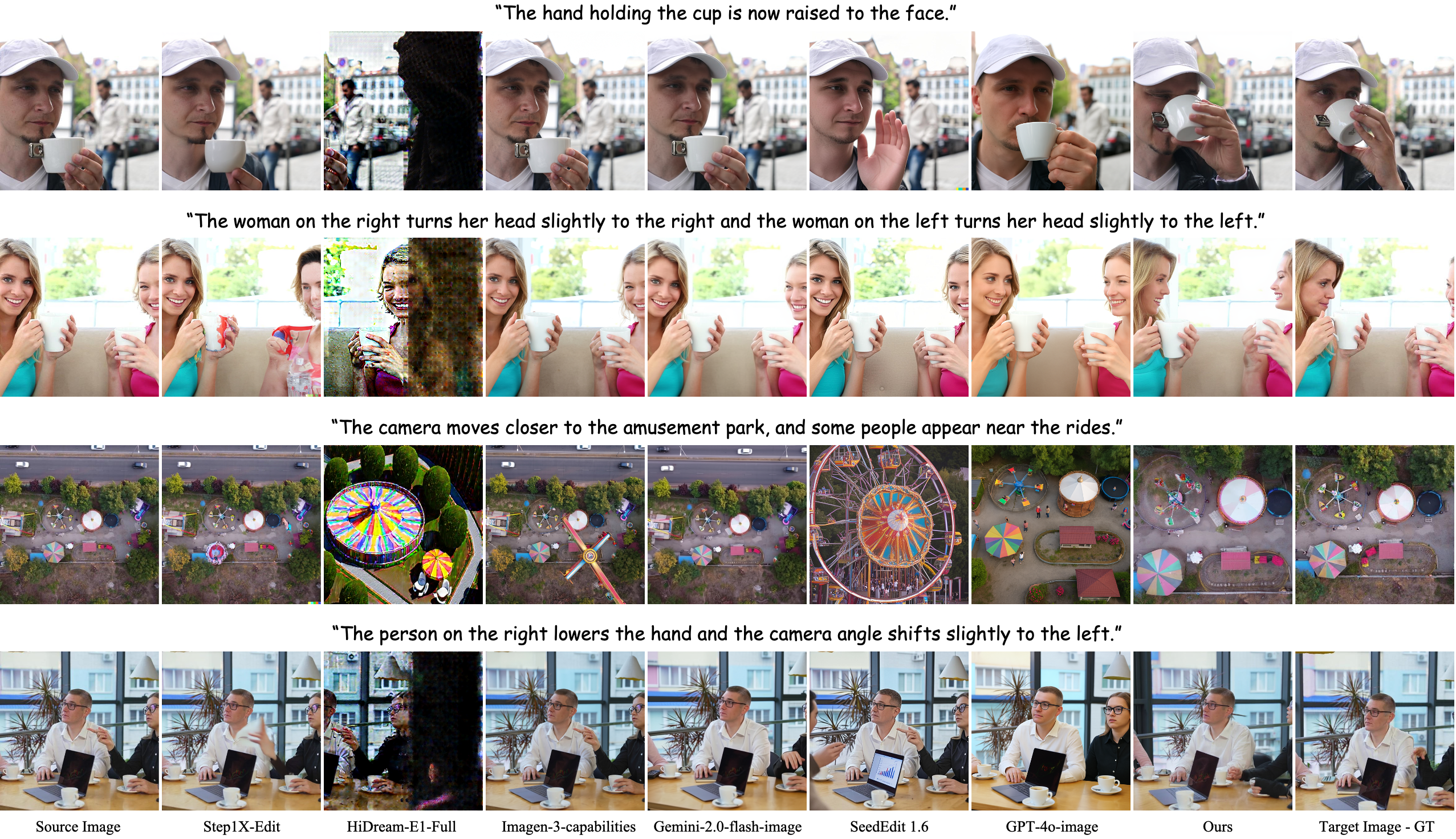}
\caption{\textbf{Qualitative comparison of \model and other methods from the industry on \benchmark.} GPT-4o-Image and \model demonstrate strong instruction-following quality while GPT-4o-Image struggles to preserve the identity and appearance in the source image. On the other hand, Gemini-2.0-flash-image and SeedEdit 1.6 preserve ID well but cannot consistently follow the motion editing instructions.
}
\label{fig:comp_1}
\vspace{-10pt}
\end{figure}

\noindent\textbf{Qualitative Results.}\label{sec:qualitative}
We present qualitative comparisons on the \benchmark in Figure~\ref{fig:comp_1}. Both GPT-4o-Image~\citep{openai2025introducing4o} and \model exhibit strong instruction-following capabilities; however, GPT-4o-Image often struggles to maintain the identity and appearance of subjects in the source image. In contrast, Gemini-2.0-flash-image~\citep{deepmind_gemini_flash} and SeedEdit-1.6~\citep{shi2024seededit} are more effective at preserving identity, but frequently fail to accurately execute the motion-editing instructions. Additional qualitative results are provided in the appendix.

\begin{figure}[ht]
    \centering
    \includegraphics[width=0.7\linewidth]{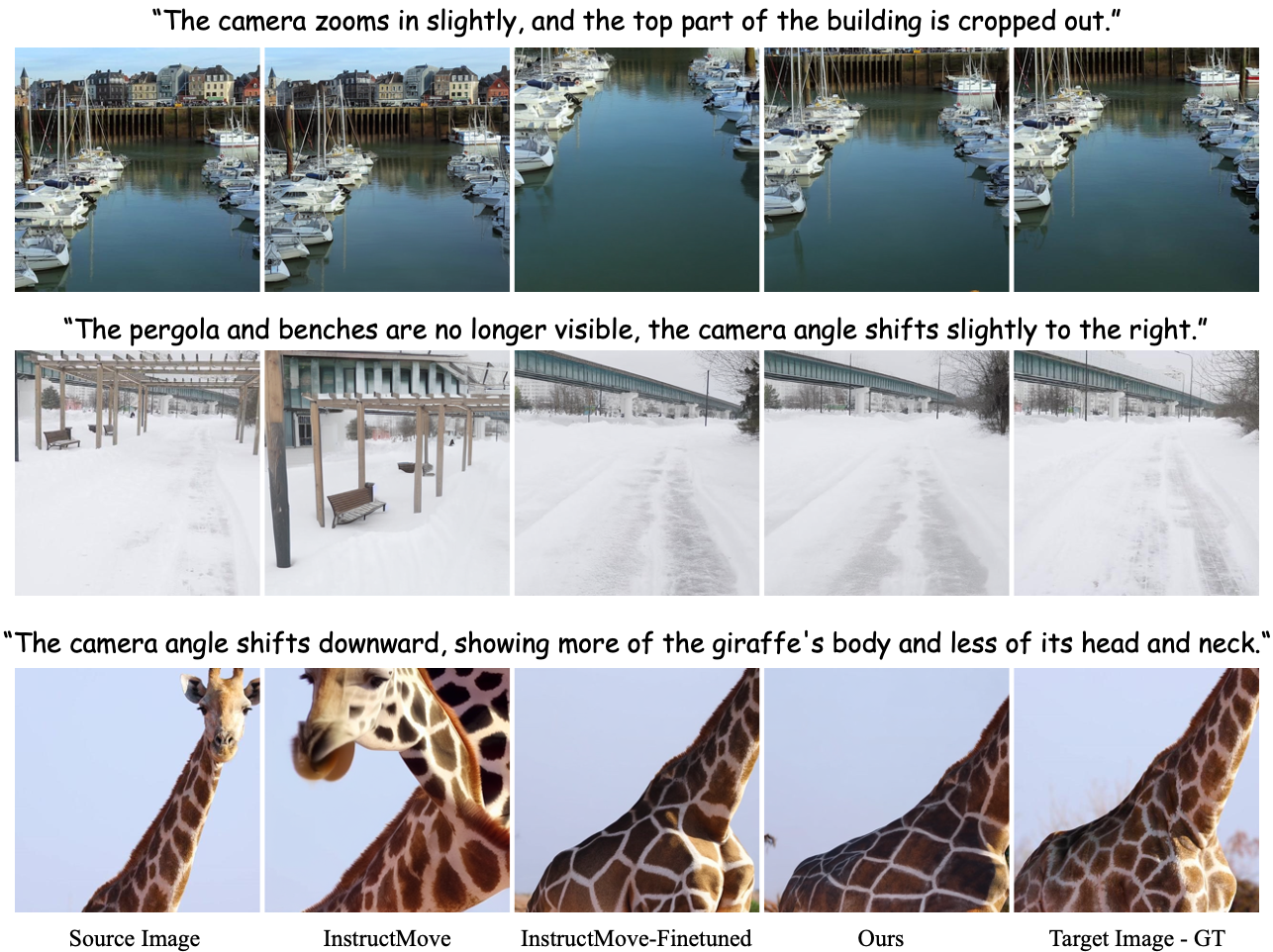}
    \caption{\textbf{Visualization of ablation analysis.} InstrucMove-Finetuned denotes we fine-tuned InstructMove~\citep{cao2024instruction} on the training set of \dataset. After such fine-tuning, the instruction-following quality for camera motion editing is significantly improved.}
    \label{fig:ablation}
    \vspace{-10pt}
\end{figure}

\noindent\textbf{Ablation Study.}\label{sec:ablation_study}
To validate the effectiveness of \dataset, we fine-tune OminiControl~\citep{tan2024ominicontrol} and InstructMove~\citep{cao2024instruction} on our training set and report the performance in Table~\ref{table:1}. Both models exhibit notable gains across key metrics after fine-tuning. Qualitative results are provided in Figure~\ref{fig:ablation}, demonstrating that models trained on \dataset achieve substantially better instruction-following ability, particularly for non-rigid motion edits.

\subsection{Limitations}\label{sec:limitation}
Since the training and evaluation data in \dataset are collected from both generated and real-world videos featuring motion changes, variations in object stylization, and relighting conditions are limited. Expanding the dataset with additional data from simulators or rendering engines would significantly increase its diversity and scale.

\section{Potential Usage and Future}\label{sec:future}
\noindent\textbf{Multi-Reference Editing and Image Reasoning.} As visualized in Figure~\ref {fig:method}, we sampled multiple images in a single video and performed the instruction labeling separately for each pair of images. It is possible to use our training set for multi-reference image editing, which is an underexplored domain. It's also possible to create models for consecutive motion editing with our dataset, with a reasoning model conditioned on previous editing instructions, source images, and corresponding image captions. We leave these tasks to the community for further exploration.

\noindent\textbf{Addressing Key Challenges.} As mentioned in Section~\ref{sec:exp}, existing methods face significant limitations, particularly in identity preservation and instruction adherence. To explore viable solutions for improving identity-aware editing, we propose two strategies: (1) \textit{Identity Embedding Consistency}. Incorporate precomputed identity embeddings by leveraging frozen encoders—such as ArcFace for human faces or DINO for general objects—and apply a consistency loss between the source and generated images during fine-tuning. This encourages the model to retain critical identity features. (2) \textit{Multi-Stage Refinement}. Generate a coarse motion-editing output in the first stage, followed by identity-preserving refinement through a second-stage module, such as an image fusion network or denoising model. This process can be enhanced using region-specific attention or cross-attention mechanisms aligned with the original image features.
To improve instruction-following, especially for non-rigid camera motions, we explore two complementary approaches: (1) \textit{Text-to-Motion Embedding with Structural Guidance}. Develop a motion grounding module that maps natural language instructions to spatial motion vectors (e.g., using the Pl{\"u}cker embedding~\citep{sitzmann2021light}), which can be integrated into the diffusion backbone. This provides explicit motion cues, similar to strategies in MotionCtrl~\citep{wang2024motionctrl}, CameraCtrl~\citep{he2024cameractrl}, and CameraCtrl II~\citep{he2025cameractrl}. (2) \textit{Instruction Perturbation Training}. Augment the training set with semantically equivalent instruction variants (e.g., “move camera to the left” vs. “shift view leftward”) to improve robustness and generalization in instruction understanding.

\section{Conclusion}\label{sec:conclusion}

In this work, we introduce \framework, a new framework for expressive image editing with non-rigid motions, bridging the gap between traditional instruction-based editing and dynamic, motion-centric transformations. To this end, we propose \dataset, a large-scale dataset designed to capture a wide spectrum of non-rigid editing scenarios, including camera movements, object deformations, human articulations, and human-object interactions. 
Extensive experiments on \benchmark demonstrate that new challenges that are not addressed in previous datasets exist. Together with \model, \framework provides a comprehensive solution for advancing instruction-driven image editing toward more realistic, expressive, and controllable scenarios.

\noindent\textbf{Social Impact Statement.}\label{sec:impact}
Our work aims to improve instruction-based image editing from a technical perspective and is not intended for misuse, such as forgery. Therefore, synthesized images should clearly indicate their artificial nature. This work is for research purposes only, and the usage of our dataset should strictly follow the rules specified in the license and guidelines.

\clearpage

\beginappendix

\appendix

\section{Open-Source}\label{sec:release}
The dataset, benchmark, code, and models are released in the following links:
\begin{itemize}
    \item Training Data Preview Subset: \\ \url{https://huggingface.co/datasets/ByteDance-Seed/BM-6M-Demo}
    \item Full Training Data: \\ \url{https://huggingface.co/datasets/ByteDance-Seed/BM-6M}
    \item Evaluation Benchmark:  \\\url{https://huggingface.co/datasets/ByteDance-Seed/BM-Bench}
    \item Evaluation Code and Instructions:  \\\url{https://github.com/ByteDance-Seed/BM-code/tree/main/ByteMorph-Eval}
    \item Baseline Model Train and Inference Code:  \\\url{https://github.com/ByteDance-Seed/BM-code/}
    \item Baseline Model Weight:  \\\url{https://huggingface.co/ByteDance-Seed/BM-Model}
    \item Project Page:
    \\\url{https://boese0601.github.io/bytemorph}
    \item Online Demo for Baseline Model:
    \\\url{https://huggingface.co/spaces/Boese0601/ByteMorph-Demo}
    
\end{itemize}

\dataset is released under CC0-1.0 Creative Commons Zero v1.0 Universal License, and the baseline model \model, including code and weights, is released under FLUX.1-dev Non-Commercial License.

\section{Details of Human Preference Study}\label{sec:user_study}

\begin{figure}[ht]
\centering
\includegraphics[width=\linewidth]{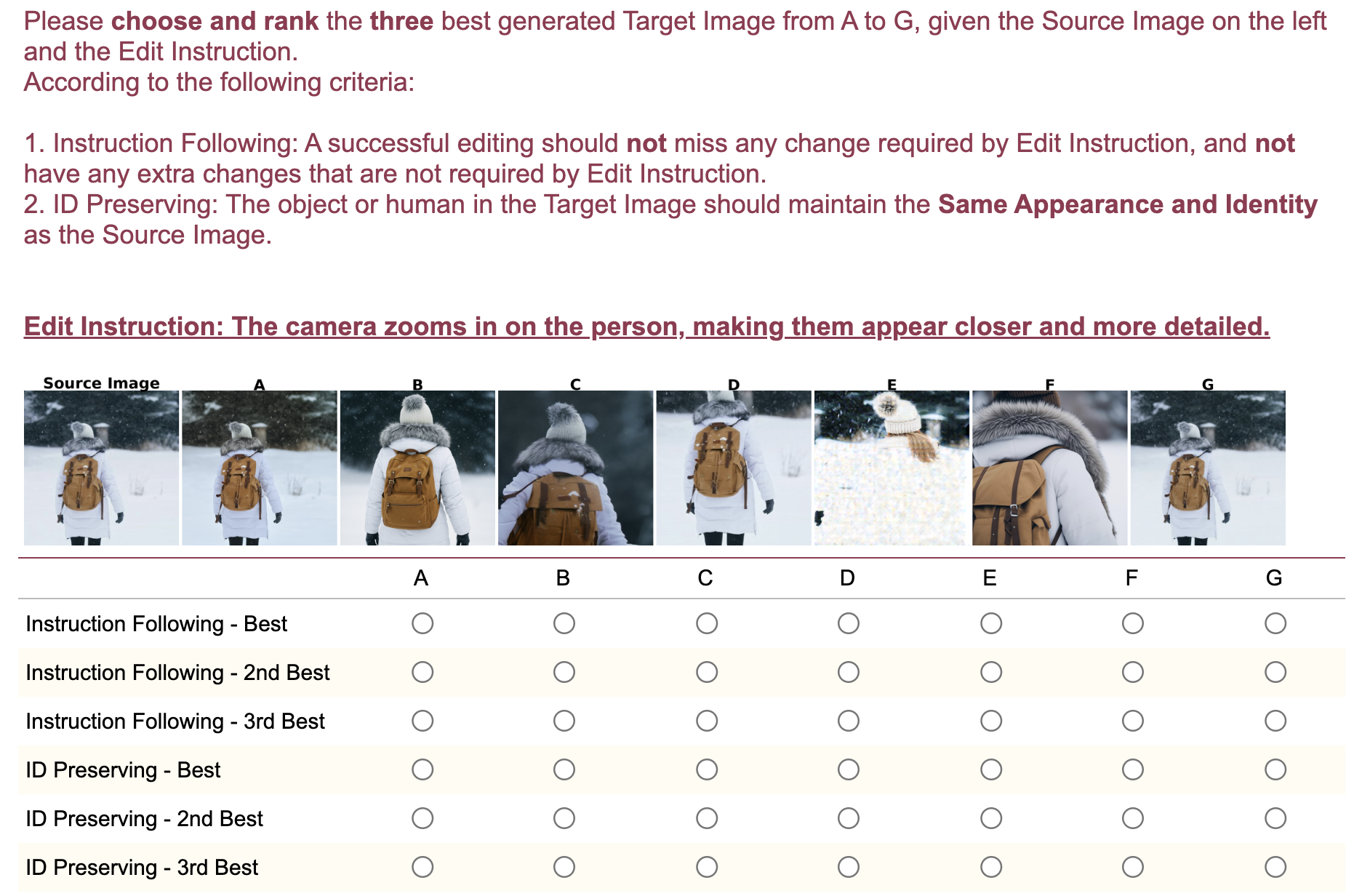}

\caption{\textbf{A screenshot of one example in our human preference study.} The user may leave the choices blank if there are not enough methods that provide satisfying results, and the user may choose multiple methods with the same ranking.
}
\label{fig:user_study_screenshot}
\end{figure}

We use \textbf{Prolific}, an online platform designed to connect academic researchers with user research participants for human-level performance evaluation. The participants are English-speaking lay people verified by this platform without prior knowledge of computer vision, and they are paid 15 USD/hr for their labor. We randomly chose ten examples from each editing category in our benchmark (50 examples in total). The results generated from different methods are anonymized and randomly sorted in each example. We then asked 40 candidates to choose and rank the \textbf{top three best} image editing results from the anonymized methods, for Instruction-Following quality and ID-Preserving ability. We collect their answers and rescale the results back to 0-100, \textit{e.g.}, the First choice = 100, the Second choice = 67, the third choice = 33, and others = 0. The participant may leave the choices blank if there are not enough methods that provide satisfying results, and the participant may choose multiple methods with the same ranking (still up to three in total). A screenshot of our human preference study is presented in Figure.~\ref{fig:user_study_screenshot}.

\section{Details on Motion Caption and Video Generation}\label{sec:add_caption}
\begin{table*}[!t]
\centering
\resizebox{\linewidth}{!}{
\begin{tabular}{>{\raggedright\arraybackslash}m{4cm}>{\raggedright\arraybackslash}m{10cm}}
\hline
 & \multicolumn{1}{c}{\textbf{Message}} \\
\hline
\textbf{System Prompt} & You are an annotator for image editing. You will be given an input image. You need to imagine a similar image editing prompt as the example image editing operation, which is suitable for the given image. The imagined editing prompt should describe at least one of these five categories:

1) The human in the given image has motion changes. 2) The object in the given image has motion changes. 3) The camera position used to take the given image zooms in or zooms out. 4) The camera position used to take the given image moves to the left or right or up or down. 5) The object or the human in the given image interacts with each other. 
The editing prompt should describe one of the five above categories.
 
Here is the example editing operation:
\{example\}

Provide your response in a JSON format as such:
\{"motion\_caption": "xxx"\}

Do not output anything else. \\
\hline
\textbf{Human Prompt} & \{base64\_source\_image\_string\} + \{example\}  \\
\hline
\textbf{Output Example} & \{"motion\_caption": "The camera angle shifts to the right and the girl on the left opens her eyes."\}  \\
\hline

\end{tabular}
}
\caption{An example of Motion Caption. The system prompt describes the VLM's responsibility as an annotator for image editing and defines the Motion Caption task. The human prompt provides the actual input to the VLM from the user. The source image is encoded as a 64-byte string according to GPT-4o~\cite{openai2024chatgpt4o} API instructions.  "example" denotes a randomly selected prompt from the template library. The output caption is saved as a JSON dictionary.}

\label{tab:vlm}
\end{table*}

In Table.~\ref{tab:vlm}, we provide an example for Motion Caption mentioned in Section.~\ref{sec:caption}.

\begin{table*}[!t]
\centering
\resizebox{\linewidth}{!}{
\begin{tabular}{>{\raggedright\arraybackslash}m{4cm}>{\raggedright\arraybackslash}m{10cm}}
\hline
 & \multicolumn{1}{c}{\textbf{Message}} \\
\hline
\textbf{System Prompt - 1} & You are given a pair of video frames. You need to observe the differences between the two frames. 
Then, summarize the changes happened between the frames in one sentence. 
Do not use past tense. Try to mention all the changes, including the motions of human body or object, the facial expressions of human, and those changes caused by camera movement. 

If there are objects and humans interact with each other, explicitly describe the interaction as well. 

Only reply the changes. 
Do not reply anything else.
\\
\hline
\textbf{Human Prompt - 1} & \{base64\_source\_image\_string\} + \{base64\_target\_image\_string\}  \\
\hline
\textbf{Output Example - 1} & \{"edit\_prompt": "The camera zooms out, several people appear on the poolside area, and a potted plant is added near the pool."
\}  \\
\hline
\hline
\textbf{System Prompt - 2} & You are given an image. You need to describe the content in this image.
Do not use the past tense. Try to mention all the subjects in this frame, including human or object.
Do not output anything else. Directly describe the image, Do not use "This image ..." or "This frame ..."
\\
\hline
\textbf{Human Prompt - 2} & \{base64\_source\_image\_string\}\\
\hline
\textbf{Output Example - 2} & \{"src\_img\_caption": "A large outdoor swimming pool with clear blue water is surrounded by a paved deck area. Multiple black lounge chairs are arranged in rows along the sides of the pool. A small building with a door and a potted plant is visible in the corner. A metal pool ladder is positioned at the edge of the pool. In the background, a cityscape with buildings and towers is visible under a partly cloudy sky."
\}  \\
\hline
\hline
\textbf{System Prompt - 3} & You are given an image. You need to describe the content in this image.
Do not use the past tense. Try to mention all the subjects in this frame, including human or object.
Do not output anything else. Directly describe the image, Do not use "This image ..." or "This frame ..."
\\
\hline
\textbf{Human Prompt - 3} & \{base64\_target\_image\_string\}\\
\hline
\textbf{Output Example - 3} & \{"tgt\_img\_caption": "A large outdoor swimming pool is surrounded by a paved area with lounge chairs neatly arranged along one side. Several women are engaged in a group exercise session on the paved area near the pool, wearing athletic clothing in various colors such as red, pink, black, and neon green. A small potted plant is placed near the pool, and the background includes a view of distant buildings and a clear sky with scattered clouds."
\}  \\
\hline
\end{tabular}
}
\caption{An example of Image-Instruction Pair Creation with \textbf{three} separate steps (denoted as \textbf{- 1}, \textbf{- 2}, and \textbf{- 3}. In the first step, the system prompt describes the VLM's responsibility as an annotator for observing the difference between source and target images, and summarizing the editing instructions. The human prompt provides the actual input to the VLM from the user. Both source and target images are encoded as 64-byte strings according to GPT-4o~\cite{openai2024chatgpt4o} API instructions. The output editing instruction is saved as a JSON dictionary. \\ In the second and third steps, the system prompt describes the VLM's responsibility as an annotator for describing the source or target image content. The human prompt provides the 64-byte encoded image string to the VLM from the user. The output caption is saved as a JSON dictionary.}\label{tab:vlm_recaption}
\vspace{-4pt}

\end{table*}

In Table.~\ref{tab:vlm_recaption}, we provide an example for Image-Instruction Pair Creation, mentioned in Section.~\ref{sec:image_instruction_pair_creation}.

\section{Details on \benchmark evaluation metrics}\label{sec:add_metrics}
\noindent \textbf{CLIP metrics.} CLIP-SIM$_{img}$ measures the cosine similarity between CLIP embedding of source image and generated image as shown in Eq.~1, CLIP-SIM$_{txt}$ evaluates the similarity between generated image and text caption of target image (Eq.~2), and CLIP-D$_{txt}$ captures the directional alignment between CLIP distance between text embedding from source to target and image from source to generation, as defined in Eq.~3.
Our proposed metric CLIP-D$_{img}$ is defined in Eq.~4. 
\begin{align}
\text{CLIP-SIM}_{\text{img}} &= \cos\left( \mathbf{C}(I_{\text{gen}}),\; \mathbf{C}(I_{\text{src}}) \right) \tag{Eq. 1} \\
\text{CLIP-SIM}_{\text{txt}} &= \cos\left( \mathbf{C}(I_{\text{gen}}),\; \mathbf{C}(T_{\text{tgt}}) \right) \tag{Eq. 2} \\
\text{CLIP-D}_{\text{txt}} &= \cos\left( \mathbf{C}(I_{\text{gen}}) - \mathbf{C}(I_{\text{src}}),\; \mathbf{C}(T_{\text{tgt}}) - \mathbf{C}(T_{\text{src}}) \right) \tag{Eq. 3} \\
\text{CLIP-D}_{\text{img}} &= \cos\left( \mathbf{C}(I_{\text{gen}}) - \mathbf{C}(I_{\text{src}}),\; \mathbf{C}(I_{\text{tgt}}) - \mathbf{C}(I_{\text{src}}) \right) \tag{Eq. 4}
\end{align}
where $\mathbf{C}(\cdot)$ stands for CLIP encoder.

\noindent \textbf{VLM-Score.} In Table.~\ref{tab:vlm_score}, we provide the detailed steps for VLM-Score. evaluation with Claude-3.7-Sonne~\cite{claude37sonnet}. 

\begin{table*}[!t]
\centering
\resizebox{\linewidth}{!}{
\begin{tabular}{>{\raggedright\arraybackslash}m{4cm}>{\raggedright\arraybackslash}m{10cm}}
\hline
 & \multicolumn{1}{c}{\textbf{Message}} \\
\hline
\textbf{System Prompt} & 
    You are an evaluator for image editing. You will be given a pair of images before and after editing as well as an editing instruction.
    You need to rate the editing result with a score between 0 to 100.
    
    A successful editing should not miss any change required by editing instruction.
    A successful editing should not have any extra changes that are not required by editing instruction. 
    The second image should have minimum change to reflect the changes made with editing instruction.
    Be strict about the changes made between two images.
    
    Give the final response in a json format as such:
    \{"Score": xx\}
    
    Do not output anything else. \\
\hline
\textbf{Human Prompt} &\{base64\_source\_image\_string\} + \{base64\_target\_image\_string\} + \{editing\_instruction\}  \\
\hline
\textbf{Output Example} & \{"Score": 90\}  \\
\hline

\end{tabular}
}
\caption{An example of VLM-Score Evaluation. The system prompt describes the VLM's responsibility as an evaluator and defines the task. The human prompt provides the actual input to the VLM from the user. Both source and target images are encoded as 64-byte strings according to Claude-3.7-Sonnet~\cite{claude37sonnet} API instructions. The output score is saved as a JSON dictionary.}\label{tab:vlm_score}
\end{table*}

\section{Additional Visualization}\label{sec:add_vis}
\begin{figure}[!ht]
\centering
\includegraphics[width=\linewidth]{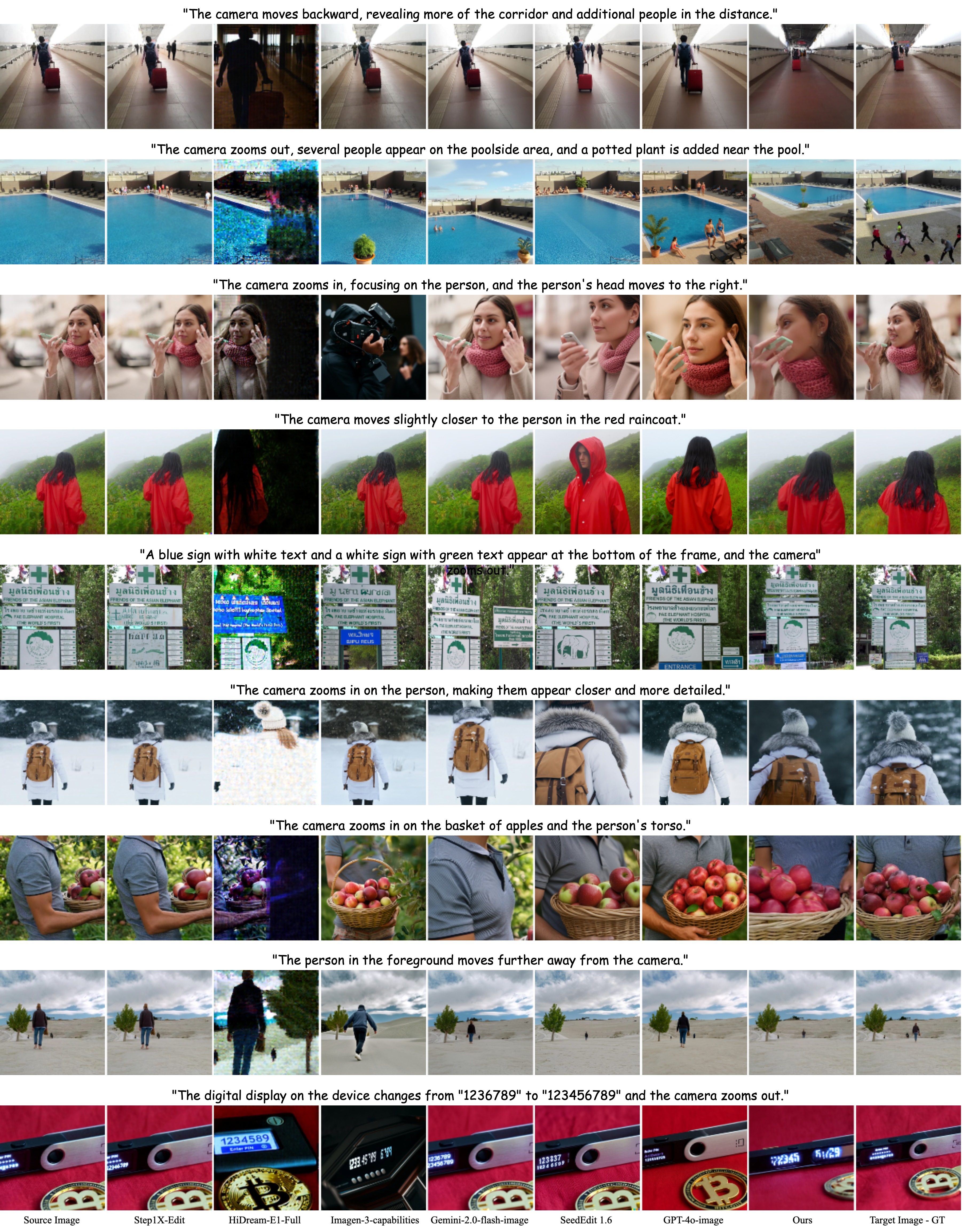}
\caption{\textbf{Visualized comparison between methods from the industry on \benchmark.} Editing Category: Camera Zoom.
}

\label{fig:supp0}
\end{figure}

\begin{figure}[!ht]
\centering
\includegraphics[width=\linewidth]{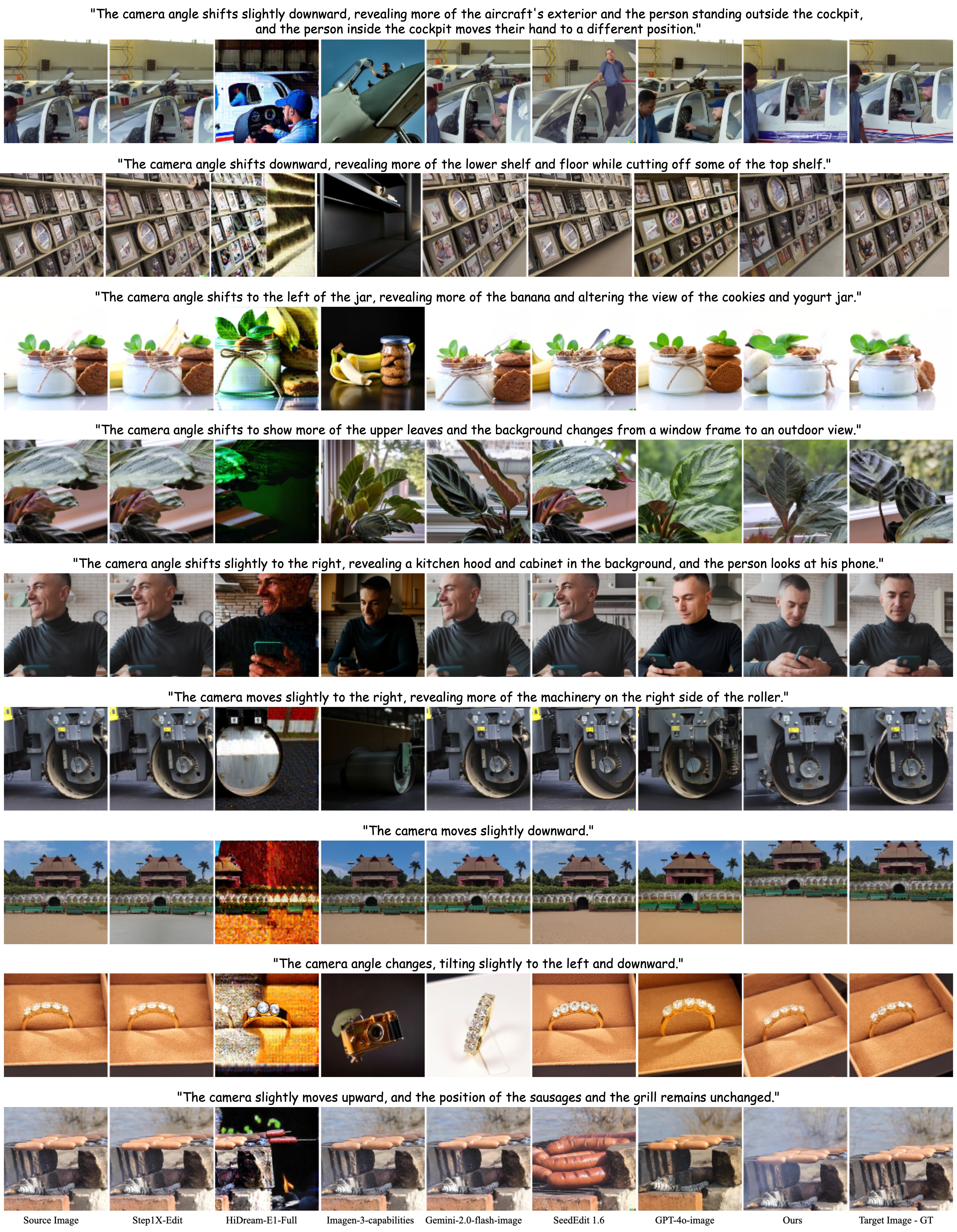}

\caption{\textbf{Visualized comparison between methods from the industry on \benchmark.} Editing Category: Camera Motion.
}

\label{fig:supp1}
\end{figure}

\begin{figure}[!ht]
\centering
\includegraphics[width=\linewidth]{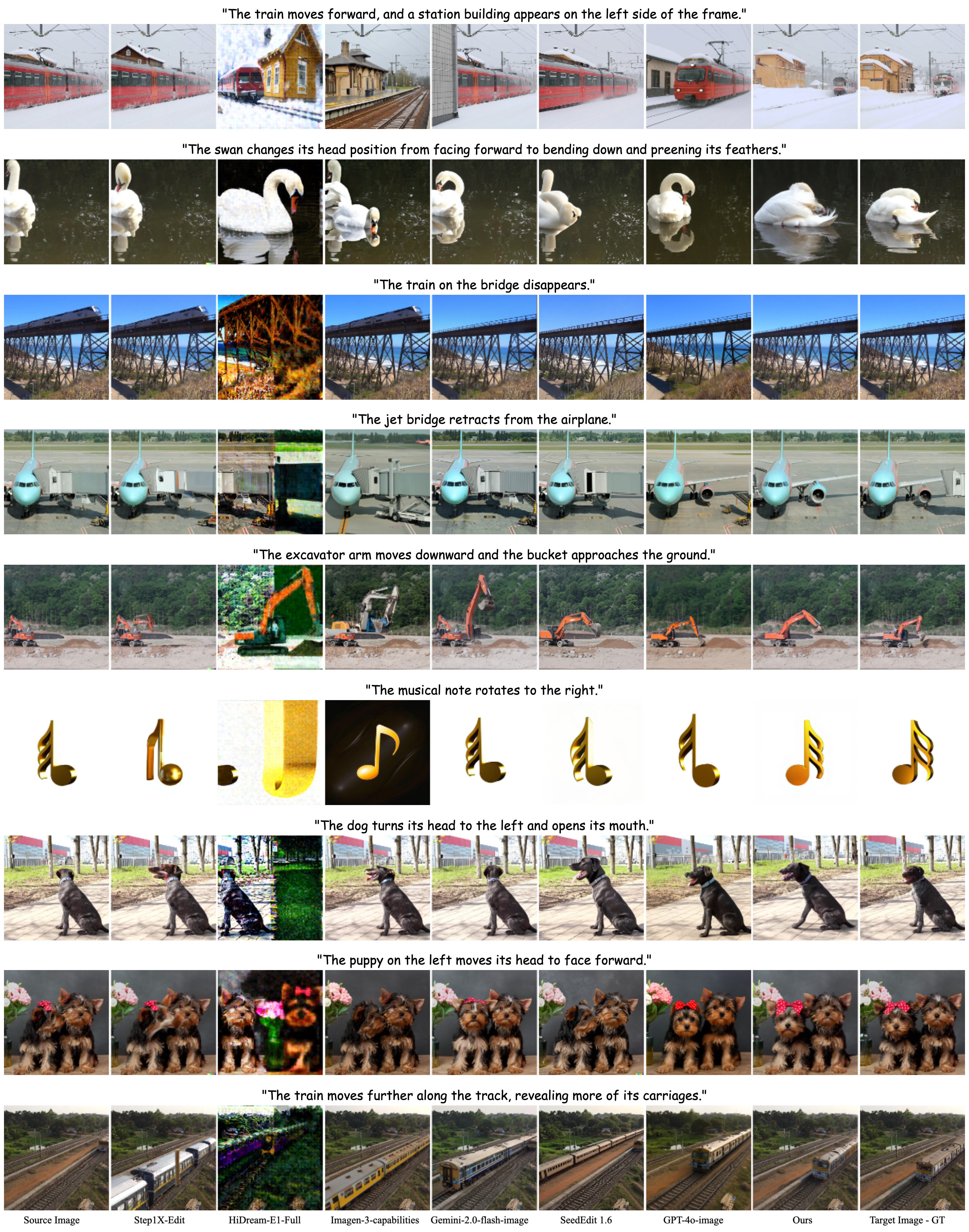}

\caption{\textbf{Visualized comparison between methods from the industry on \benchmark.} Editing Category: Object Motion.
}

\label{fig:supp2}
\end{figure}

\begin{figure}[!ht]
\centering
\includegraphics[width=\linewidth]{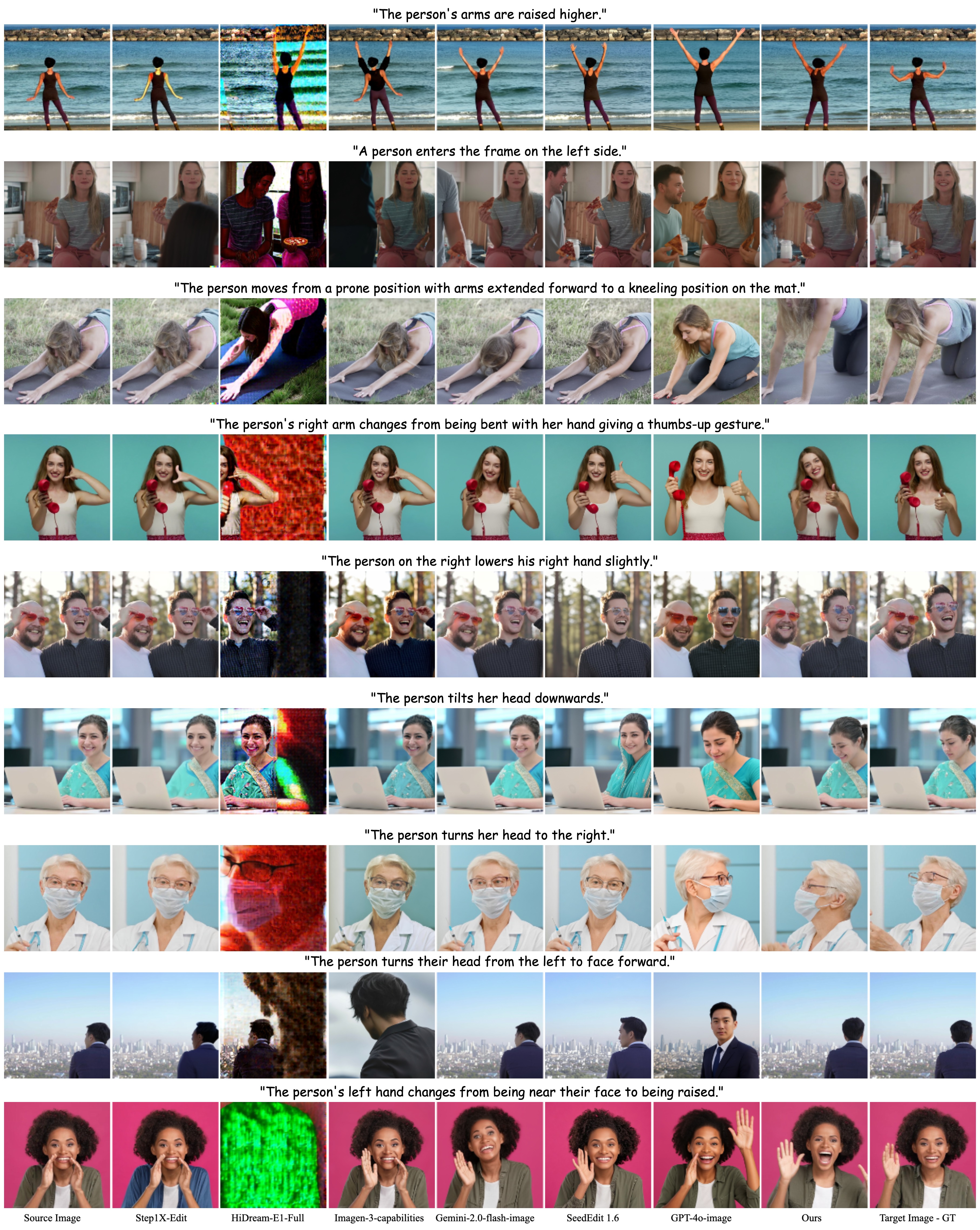}

\caption{\textbf{Visualized comparison between methods from the industry on \benchmark.} Editing Category: Human Motion.
}

\label{fig:supp3}
\end{figure}

\begin{figure}[!ht]
\centering
\includegraphics[width=\linewidth]{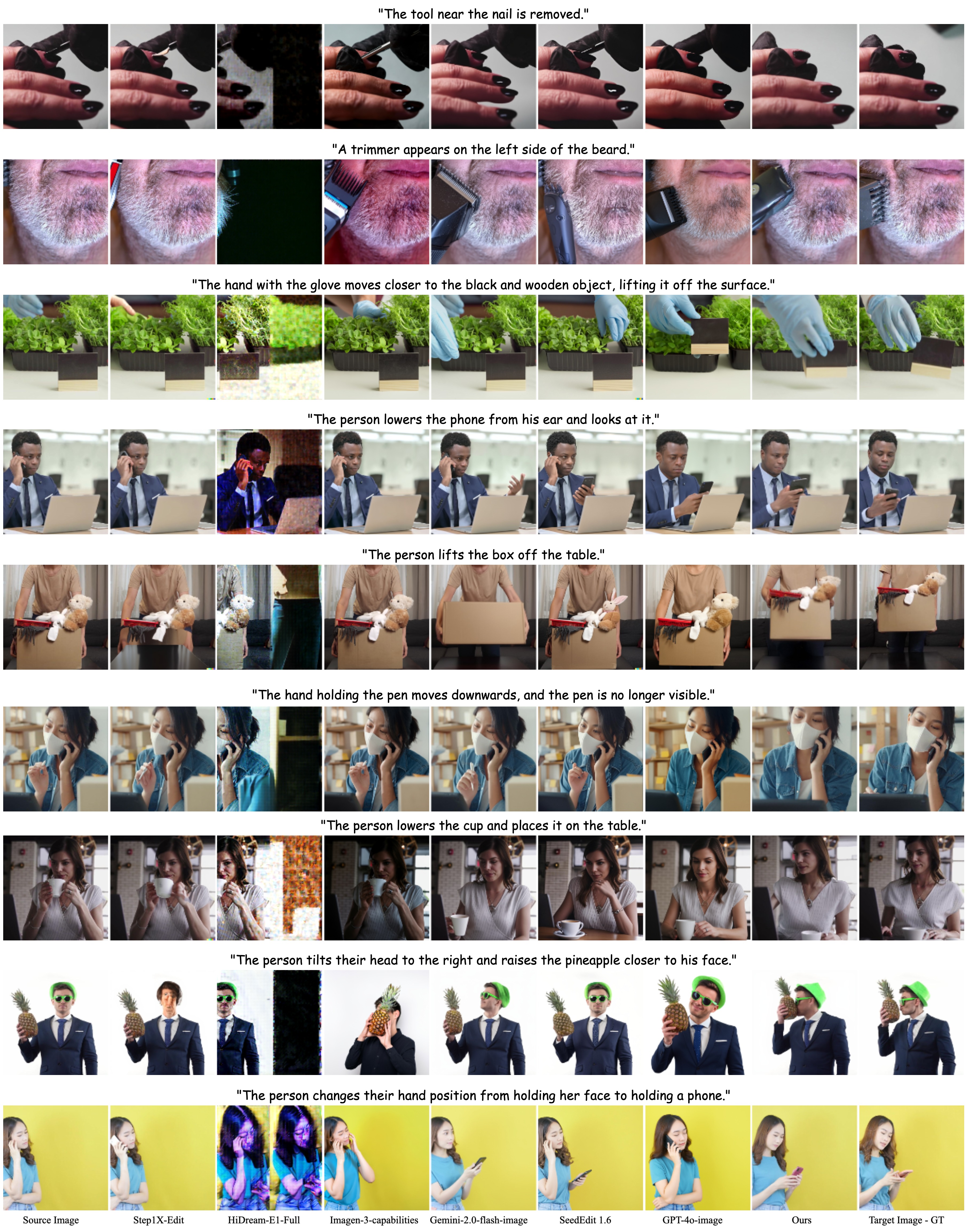}

\caption{\textbf{Visualized comparison between methods from the industry on \benchmark.} Editing Category: Human-Object Interaction.
}

\label{fig:supp4}
\end{figure}

We provide additional visualized comparison between commercial methods~\cite{liu2025step1x-edit,hidream_e1,imagen3,shi2024seededit,deepmind_gemini_flash,openai2025introducing4o} and our proposed \model in Figure.~\ref{fig:supp0}, Figure.~\ref{fig:supp1}, Figure.~\ref{fig:supp2}, Figure.~\ref{fig:supp3}, and Figure.~\ref{fig:supp4}. GPT-4o-image and \model both demonstrated strong instruction-following ability, while GPT-4o-image performed worse than the Gemini-2.0-flash-image, SeedEdit 1.6, and \model in identity preservation.

{\small
    \clearpage
    \bibliographystyle{ieee}
    \bibliography{ref}
}

\newpage

\end{document}